\documentclass[journal]{IEEEtran}
\usepackage{cite}
\usepackage{amsmath,amssymb,amsfonts}
\usepackage{algorithmic}[1]
\usepackage{graphicx}
\usepackage{textcomp}
\usepackage{url}
\usepackage{hyperref}
\usepackage{multirow}
\usepackage[ruled,vlined]{algorithm2e}
 \usepackage{float}
\usepackage[framemethod=default]{mdframed}
\newmdenv[
  topline=true,
  bottomline=true,
  skipabove=6pt,
  skipbelow=6pt,
  linewidth=0.8pt,
  innerleftmargin=6pt,
  innerrightmargin=6pt,
]{promptbox}
\usepackage{makecell}

\usepackage{bm}

\def\BibTeX{{\rm B\kern-.05em{\sc i\kern-.025em b}\kern-.08em
    T\kern-.1667em\lower.7ex\hbox{E}\kern-.125emX}}

\begin{document}

\title{Think Inside the Chunk: RegulaRAG for Regulation-Compliant Scenario Generation using LLMs: A Case Study of UN Regulation No. 152}

\author{Vahid~Zolfaghari,~\IEEEmembership{IEEE},
        Nenad~Petrovic,
        Andr\'e~Schamschurko,
        and~Alois~Knoll,~\IEEEmembership{Fellow,~IEEE}%
\thanks{V. Zolfaghari, N. Petrovic, A. Schamschurko, and A. Knoll are with the Chair of Robotics, Artificial Intelligence (AI) and Embedded Systems, Technical University of Munich, Munich, Germany (e-mail: v.zolfaghari@tum.de; nenad.petrovic@tum.de; andre.schamschurko@tum.de; k@tum.de).}%
\thanks{Corresponding author: Vahid Zolfaghari (e-mail: v.zolfaghari@tum.de).}%
\thanks{This research was funded by the Federal Ministry of Research, Technology and Space of Germany (BMFTR) as part of the CeCaS project, FKZ: 16ME0800K.}}

\markboth
{Zolfaghari \MakeLowercase{\textit{et al.}}: Think Inside the Chunk: RegulaRAG for Regulation-Compliant Scenario Generation using LLMs: A Case Study of UN Regulation No. 152}
{Zolfaghari \MakeLowercase{\textit{et al.}}: Think Inside the Chunk: RegulaRAG for Regulation-Compliant Scenario Generation using LLMs: A Case Study of UN Regulation No. 152}

\maketitle

\begin{abstract}
Generating regulation-compliant test scenarios is essential for validating safety-critical automotive systems, yet Large Language Models (LLMs) struggle to ground outputs in long, hierarchical standards. We present \textbf{RegulaRAG}, a Retrieval-Augmented Generation (RAG) pipeline that couples SmartChunking, reference-aware enrichment of paragraphs and tables via graph traversal, with Smart Retrieve \& Rerank over these enriched units. To test our system, we evaluate on a manually curated dataset covering all scenarios in UN Regulation No.~152 (AEBS). Our study comprises: (i) a three-step progressive search that identifies near-optimal retrieval parameters without exhaustive grid search; (ii) head-to-head comparisons against five baseline RAG systems; and (iii) a robustness stress test that scales the source corpus with distractor content. Outputs are evaluated using a customized penalized scoring metric.
Across all experiments, RegulaRAG achieves the highest average Meta-Score (82.99), outperforming the next-best system by 43\% (NoRAG: 57.94), while operating at 14k–25k tokens per query versus up to 500k for graph-centric baselines.
It maintains strong performance, remaining stable even as the number of regulatory sources grows, whereas competing RAG systems degrade sharply in both quality and robustness.

\end{abstract}

\begin{IEEEkeywords}
Automotive software, Large Language Models, Retrieval Augmented Generation, Standard compliance
\end{IEEEkeywords}

\section{Introduction}
\label{sec:introduction}
\IEEEPARstart{L}{arge} Language Models (LLMs) such as GPT-4o demonstrate impressive capabilities in generating human-like responses across various domains. However, LLMs face challenges such as hallucinations, outdated knowledge, and difficulty in recalling rare or highly specific information \cite{kandpalLargeLanguageModels2023,mallenWhenNotTrust2023}.
To overcome these limitations, Retrieval-Augmented Generation (RAG) has emerged as a promising approach by integrating external documents during inference to enhance accuracy, traceability, and grounding \cite{leeLatentRetrievalWeakly2019}.
\\
In the context of automotive software development, particularly for Advanced Driver Assistance Systems (ADAS) and Autonomous Driving Systems (ADS), generating precise, regulation-compliant test scenarios is critical. Traditional manual processes for interpreting and validating against regulatory standards like UN Regulation No. 152 are time-consuming, error-prone, and costly. LLMs have potential to automate this effort, but without RAG, they often struggle to identify the correct numerical parameters, differentiate test conditions (e.g., laden vs. unladen), and manage long documents exceeding context limits. Furthermore, directly inputting the full regulation to an LLM results in high token usage and associated costs. Although recent large language models (LLMs) such as Claude provide extended context windows of up to 100k tokens ~\cite{Introducing100KContext}, simply feeding full-length automotive documents (e.g., requirements or specifications) into the prompt is rarely feasible due to confidentiality and intellectual property concerns. Moreover, long-context models often struggle with effective utilization of all input tokens, suffering from issues such as context dilution or the so-called "lost in the middle" effect~\cite{liuLostMiddleHow2023}.
\\
To address these challenges, we introduce \textbf{RegulaRAG}, a smart two-stage Retrieval-Augmented Generation pipeline tailored for regulation-compliant scenario generation. RegulaRAG employs a novel \textit{SmartChunking} strategy to preprocess PDF-based standards, identify hierarchical paragraph structures and nested references, and enrich document chunks via a graph-based traversal. This ensures that retrieved content includes all semantically linked paragraphs and tables, while maintaining a compact token footprint. Our \textit{Smart Retrieve and Rerank} module performs query-aware retrieval over these reference-enriched chunks, allowing the LLM to generate accurate, traceable, and
standards-compliant scenarios. This method outperforms traditional retrieval pipelines that operate on uninformed or semantically shallow chunks. An example of such traditional methods is provided in~\cite{zolfaghariAdoptingRAGLLMAided2024}, where a simple Retrieve-and-Rerank strategy was applied for question answering over automotive documents to support hardware and software design workflows.
The core methodological novelty of RegulaRAG lies in two algorithmic contributions: (i)~reference-aware chunk enrichment via BFS graph traversal over the regulation's explicit cross-reference structure, and (ii)~retrieval and reranking against these enriched representations to surface dispersed regulatory evidence in a single step. Bounded BFS depth, chunk deduplication, and metadata compression are \emph{engineering optimizations} that improve practical scalability on long regulations without altering the underlying algorithm; they are reported as such in Section~\ref{sys_desc}.
\\
The remainder of this paper is structured as follows: Section \ref{liter} reviews related work on LLMs for compliance and scenario generation. Section~\ref{sys_desc} presents the RegulaRAG system and the dataset. Section~\ref{exp_setup} describes the experimental setup. Section~\ref{res_and_exp} presents the experiments and results. Section \ref{sec:conclusion} concludes the paper.

\section{Literature Review}\label{liter}
\begin{figure*}[t!]
\centering
\includegraphics[width=0.5\textwidth]{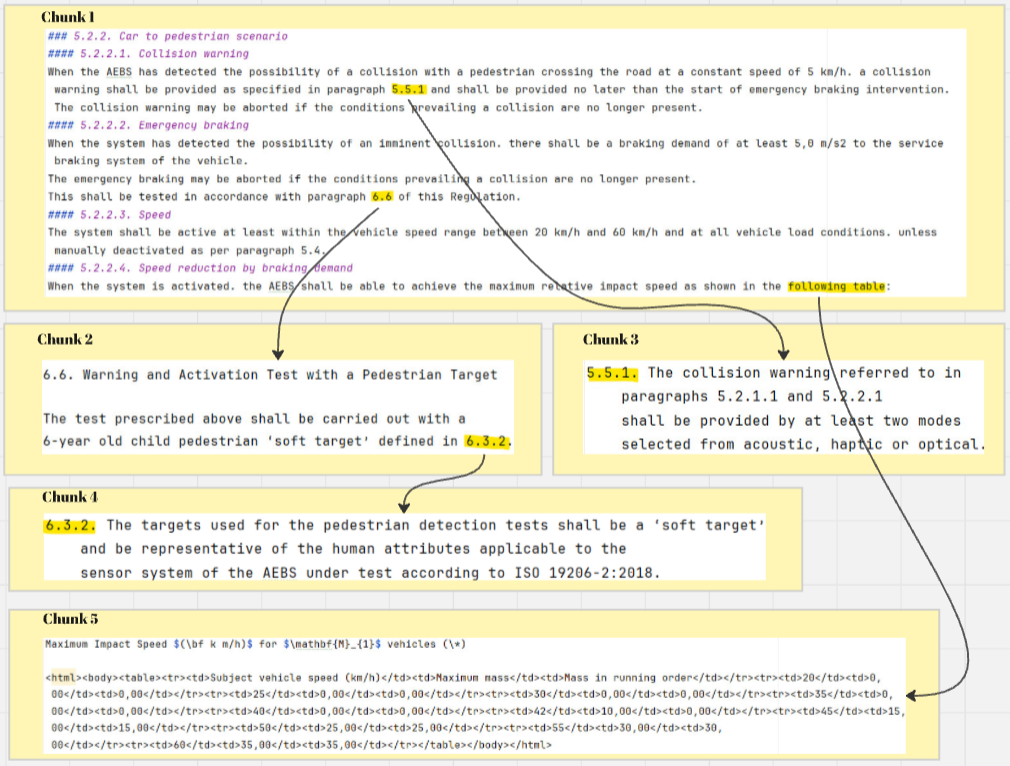}
\caption{\textbf{Example of a reference and table chain in \textit{UN Regulation No. 152}.}\label{fig:sample_chain}}
\end{figure*}

The challenge of processing complex PDF documents, a common format for technical specifications in the automotive industry, has been addressed by several researchers. For example, \cite{liuOptimizingRAGTechniques2024a} propose a method that addresses key challenges in automotive document processing, including multi-column layouts and technical specifications. Following is the main challenges in processing these documents:  
\begin{enumerate}
    \item \textbf{Tables:} Automotive standards often contain critical tables with specifications, test results, or compliance data. These may span multiple pages in the PDF and get split into separate tables after conversion, while in fact they must be interpreted as one \cite{liuOptimizingRAGTechniques2024a}. An example of these split tables can be seen in Fig.~\ref{fig:splittable}. 
    
    \item \textbf{Domain-specific terminology:} Regulations use highly technical and legal jargon, abbreviations, and region-specific terms. Since RAG relies on similarity between queries and chunks, such variation often causes it to miss relevant content.
    
    \item \textbf{Lexical and syntactic variation:} The same requirement can appear in different formulations, typically as long, complex sentences with passive voice and nested clauses. This makes retrieval and interpretation challenging \cite{niuTVRAutomotiveSystem2025} .
    
    \item \textbf{Nested and interlinked references:} Requirements are frequently split across multiple clauses and annexes that point to one another. An example of these interlinked references is illustrated in Figure~\ref{fig:sample_chain}. Standard RAG systems often retrieve isolated chunks, missing the linked context.
\end{enumerate}

Recent research has increasingly focused on leveraging Large Language Models (LLMs) for compliance verification across regulated domains. 
Sun et al.~\cite{sunComplianceCheckingFramework2025} propose a compliance checking framework that integrates Retrieval-Augmented Generation (RAG) with an ``eventic graph'' to align business process descriptions with regulatory rules. 
Their method demonstrates the benefits of combining structured knowledge with retrieval-based augmentation, but it primarily produces compliance verdicts rather than executable artifacts. Bolton et al.~\cite{boltonDocumentRetrievalAugmented2025} introduce DRAFT, a document-retrieval-augmented fine-tuning approach for safety-critical software assessments. By fine-tuning LLMs with both standards and distractors, their method improves factual correctness and evidence handling. However, the output remains focused on assessment text and explanations rather than concrete test cases. 
LLMs and RAG are also used for detecting inconsistencies, contradictions, and conflicts in regulatory documents. Kumar and Roussinov~\cite{kumarNLPbasedRegulatoryCompliance2024} explore the use of GPT-4 for these purposes. While effective for semantic analysis at the document level, the method does not extend toward structured instantiation of compliance requirements. A complementary line of work seeks to translate regulatory clauses into executable rules. Li et al.~\cite{liCompliancetoCodeEnhancingFinancial2025} present \textit{Compliance-to-Code}, which maps financial regulations into programmatic logic to automate compliance checks.  Although this bridges textual requirements and computational enforcement, it remains abstract and disconnected from domain-specific simulation semantics. 

Agarwal et al.~\cite{agarwalRAGulatingComplianceMultiAgent2025} propose a multi-agent knowledge graph framework for regulatory Quesion and Answer, where extracted triplets are combined with RAG-based retrieval to improve factual correctness and traceability. This enhances transparency in compliance reasoning but is not tailored to generating domain-grounded artifacts such as test scenarios. 

Compared to these approaches, our work targets a different level of compliance verification. Rather than producing static classifications, narrative explanations, or direct code translations, \textit{RegulaRAG} operationalizes regulatory text into simulation-ready test scenarios. By combining SmartChunking, reference-aware enrichment, and selective retrieval, our method dynamically reconstructs scenario definitions from regulatory clauses, tables, and cross-references in a traceable manner.

A more recent line of work targets \emph{structure-aware} retrieval over complex documents.
Xu et al.~\cite{xuEquippingRetrievalAugmented2025} propose RDR2, a Retrieve--Document Route--Read pipeline in which an LLM-based router navigates a document structure tree at inference time to select or expand the most relevant heading nodes before generation.
RDR2 shows strong gains on general QA benchmarks (TriviaQA, HotpotQA, ASQA), but its structure prior is a heading hierarchy and does not address table reconstruction across page boundaries, typed legal cross-references, or normative dependencies between provisions.
Yu et al.~\cite{yuTableRAGHeterogeneous2025} propose TableRAG, which addresses the loss of table integrity in standard chunking by storing tables in a relational database and enabling SQL-driven retrieval alongside text-based search.
TableRAG is the closest prior work on the table-preservation problem, yet it targets document QA over Wikipedia-style sources and does not handle regulatory cross-references or downstream scenario synthesis.
Closer to the normative domain, Oliveira et al.~\cite{oliveiraRetrievalAugmentedGeneration2026} build a knowledge graph over Portuguese legal resolutions with typed nodes (Document, Article, Paragraph) and edges that explicitly model amendment and revocation relations (Contain, Modify, Revoke). Their system expands retrieved seed nodes to graph neighbours but discards table and image content and does not produce structured scenario artefacts.
RegulaRAG differs from all three in targeting regulation-specific \emph{scenario generation}: it reconstructs fragmented PDF tables via a rule-based heuristic, performs BFS reference-closure enrichment over the regulation's explicit cross-reference graph, and produces compliance-grounded scenario text rather than short factual answers.

Unlike knowledge-graph-based approaches, RegulaRAG does not perform general entity recognition or relation extraction. This is a deliberate design choice grounded in the properties of regulatory text: legal language in standards such as UN Regulation No.~152 relies on heavily nested clause structures, conditional syntax, and domain-specific cross-references that make reliable open-domain entity detection and edge extraction challenging. Rather than building a general-purpose KG, RegulaRAG exploits the regulation's \emph{explicit} cross-reference structure (numbered paragraph headings and direct table pointers already embedded in the document) as a lightweight, structure-preserving proxy for knowledge linkage. This avoids the LLM-heavy NER, triple extraction, and graph-expansion passes that drive the high token costs observed in KG-based pipelines (see Section~\ref{sec:scalability} for a direct comparison).

While we evaluate RegulaRAG on UN Regulation No.\,152 \cite{unionRegulationNo1522020}---which specifies the performance, testing procedures, and safety requirements for Automated Emergency Braking Systems (AEBS) in passenger vehicles---the pipeline is not specific to AEBS or to Regulation~152. In principle, the same workflow can be applied to other UN regulations that share similar structural properties (hierarchical paragraphs, annexes, and interlinked references) and can potentially be adapted to other languages with very little modification. Applying RegulaRAG to a different UN regulation primarily requires updating the prompts, as the core algorithms for chunk enrichment, de-duplication, and retrieval remain universal. This shifts compliance validation from purely textual analysis to the generation of executable test scenarios, bridging the gap between regulatory interpretation and practical system validation in safety-critical testing workflows.

In the technical standards domain, CompliAT \cite{aroraStandardsCompliantAssistiveTechnology2024} presents a structured approach for compliance checking of Assistive Technology (AT) products against relevant standards. It focuses on verifying terminological consistency with standard-defined terms, correctly classifying products, and validating product specifications against applicable requirements. However, unlike these works, which primarily aim at classification or static compliance checks, our work dynamically generates test scenarios based on regulatory documents to validate system behavior through simulation-based execution.

A parallel line of research applies LLMs to the automated generation of driving scenarios for ADS testing. 
Chat2Scenario~\cite{zhaoChat2ScenarioScenarioExtraction2024} extracts concrete simulation scenarios from naturalistic datasets using GPT-4, while TARGET~\cite{dengTARGETAutomatedScenario2023} translates textual traffic rules into a formal DSL for test generation. 
LeGEND~\cite{tangLeGENDTopDownApproach2024} derives scenarios from accident reports to enhance diversity and realism, and LEADE~\cite{tianLMMenhancedSafetyCriticalScenario2025} leverages multimodal prompting on traffic videos to reconstruct functional scenarios where ADS may fail. 
OmniTester~\cite{luMultimodalLargeLanguage2024} combines multimodal LLMs with user prompts to produce diverse and critical scenarios. 
While these works focus on creating challenging or diverse test cases to stress ADS in simulation, our objective differs: rather than inventing novel or corner-case scenarios, RegulaRAG systematically extracts \emph{regulation-compliant} scenarios directly from formal standards such as UN Regulation No.\,152, thereby prioritizing completeness, traceability, and compliance over diversity alone.
Because our method generates regulation-faithful scenarios in natural language, they can serve as inputs to downstream code-generation systems that set up simulations; for example, Lebioda et~al.~\cite{lebiodaAreRequirementsReally2025a} show that LLMs can translate abstract requirements from automotive regulations into executable \textsc{CARLA} configuration code.

In summary, while all these works demonstrate the power of LLMs in test scenario generation, our approach is unique in targeting the structured extraction of test scenarios from regulatory standards for compliance verification—bridging the gap between scenario generation and formal safety requirements in a traceable and automated manner.

\section{System Description}\label{sys_desc}
\begin{figure*}[t!]
\centering
\includegraphics[width=0.9\textwidth]{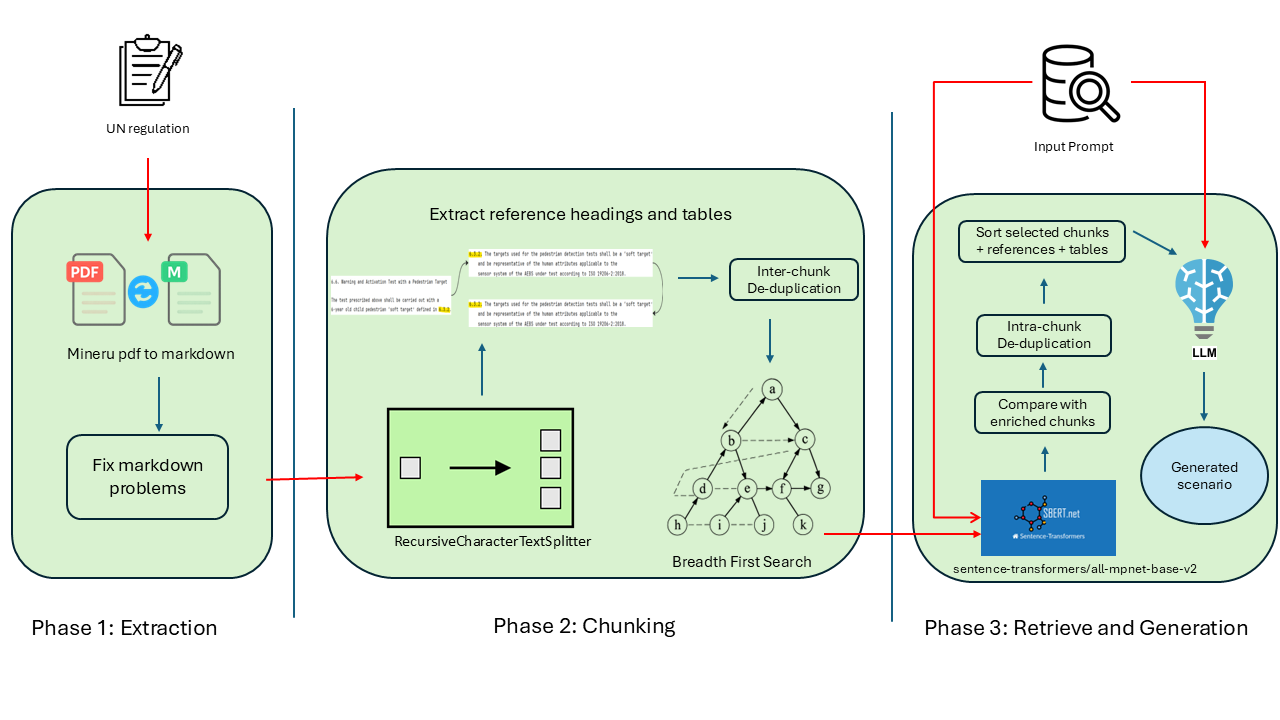}
\caption{\textbf{RegulaRAG pipeline}\label{fig:pipeline}}
\end{figure*}

The main goal of the RegulaRAG pipeline is to extract and structure the most relevant parts of lengthy regulation documents so that Large Language Models (LLMs) can generate regulation-compliant test scenarios. An overview of the full RegulaRAG pipeline can be seen in Figure~\ref{fig:pipeline}. The pipeline operates in three phases:

\paragraph{Phase 1: Extraction.}
We use the Mineru tool~\cite{wangMinerUOpenSourceSolution2024} to convert the regulation document from PDF into Markdown format and then normalize the extracted text to make it machine-usable. Concretely, we convert legal paragraph numbers (e.g., 5.2.1.4) into canonical Markdown headers, repair extraction artifacts (line breaks, hyphenation), and standardize tables while preserving the original numbering as stable IDs. The resulting structure keeps chunk boundaries aligned with the regulation's paragraph hierarchy, makes cross-references detectable for reference-aware enrichment, and provides cleaner inputs for retrieval and re-ranking.

\paragraph{Phase 2: Chunking (SmartChunking).}
Given the structured Markdown produced in Phase~1, SmartChunking first segments the text into semantically coherent base chunks using LangChain's \textsf{RecursiveCharacterTextSplitter}\footnote{\url{https://python.langchain.com/api_reference/text_splitters/character/langchain_text_splitters.character.RecursiveCharacterTextSplitter.html}}, then builds a reference-closure for each chunk by resolving cross-referenced headings and tables via BFS, so every chunk is self-contained with respect to the regulation's internal dependencies. We separate the core algorithmic steps from engineering optimizations below.

Following is the core algorithm for SmartChunking: 
\begin{enumerate}
    \item \textbf{Detect references (heading and table references).}
    For each chunk, we record three types of metadata using rule-based patterns over the normalized text:
    \begin{enumerate}
        \item \textbf{Defined Headings}: Headings explicitly defined within the chunk using Markdown syntax (lines starting with \texttt{\#}).
        \item \textbf{Referenced Headings}: Identifiers of headings (e.g., ``2.3'') mentioned in the chunk but defined elsewhere. These are resolved by locating the defining chunk and linking it.
        \item \textbf{Referenced Tables}: References to HTML-formatted tables identified using heuristics (e.g., ``the following table''). Chunks mentioning such phrases are linked to the chunk containing the actual table.
    \end{enumerate}
    Heading references are resolved by locating the chunk that defines the target heading ID.
    For table references, we apply a \textbf{rule-based heuristic}: any paragraph containing a trigger phrase such as ``the following table'' is treated as an explicit pointer to subsequent table content.
    We then collect the consecutive \texttt{<table>} HTML blocks that immediately follow the referencing paragraph in Mineru's Markdown/HTML output and group them as a single logical table entry.
    This grouping reconstructs PDF tables that were split across pages and converted into multiple adjacent HTML fragments.
    Each resolved reference is mapped to its target stable regulation ID.

    \item \textbf{Build a document reference graph.}
    We construct a directed graph where nodes represent heading/table IDs (and their defining chunks), and edges represent explicit references from one node to another. This graph encodes the regulation's cross-reference structure.

    \item \textbf{Compute transitive reference closure via BFS.}
    After metadata extraction, we expand each chunk's context by traversing the reference graph via Breadth-First Search (BFS), recursively collecting all referenced headings/tables including nested chains (e.g., a chunk referencing 6.6, which in turn references 6.3.2, as depicted in Figure~\ref{fig:sample_chain}). This produces a reference-closure set of evidence associated with the base chunk. During this stage, we apply inter-chunk deduplication to eliminate redundant content. For instance, if a chunk references two subparagraphs defined within the same chunk, we ensure that the defining chunk is not added twice. This deduplication ensures efficient context expansion and prevents excessive token usage.

    \item \textbf{Create enriched chunk representations (base + closure).}
    We form an \emph{enriched representation} for each chunk by concatenating the base chunk with the text of its reference-closure items (headings/tables). These enriched representations are used as retrieval units so that evidence dispersed across paragraphs and annexes can be retrieved together.
\end{enumerate}

To keep SmartChunking practical on long regulations, we apply several Engineering optimizations that do not change the underlying algorithm:
(i) \textbf{bounded BFS} (we cap BFS expansion to a maximum of 50 nodes per chunk, i.e., \texttt{max\_expand}$=50$, to prevent pathological expansion in densely cross-referenced regulations);
(ii) \textbf{de-duplication} (we remove repeated referenced items both within a chunk's closure and across overlapping closures to avoid redundant context);
and (iii) \textbf{metadata compression} (we store references as integer IDs/pointers instead of copying full referenced text during preprocessing, materializing text only when assembling retrieval contexts). We report these optimizations explicitly as implementation choices aimed at improving runtime and token efficiency.

\paragraph{Phase 3: Retrieval and Generation.}
Smart Retrieve and Rerank differs from standard top-$k$ retrieval in that similarity is computed between the query and each chunk's \emph{enriched} representation (base text concatenated with its reference-closure), so that a query for a test condition can match a chunk through its referenced table or sub-paragraph even when the base text alone would score poorly; the retrieved base chunks and their closure items are then assembled into the final LLM context. In the third phase, we embed all enriched chunk representations with \textsf{sentence-transformers/all-mpnet-base-v2}\footnote{\url{https://huggingface.co/sentence-transformers/all-mpnet-base-v2}} from the sentence-transformers library (v3.4.1), which maps each sentence or paragraph to a 768-dimensional dense vector. This model is based on MPNet and is widely used for semantic search and dense retrieval due to its strong performance on sentence-level similarity benchmarks. We selected this model for retrieval because it provides robust semantic matching performance while maintaining moderate computational cost, which is important for large regulatory corpora.

To compute similarity between the query and each enriched chunk, we use the cosine\_similarity function from the scikit-learn library.\footnote{\url{https://scikit-learn.org/stable/modules/generated/sklearn.metrics.pairwise.cosine_similarity.html}} Based on these scores, we retrieve the top-$k$ most relevant chunks.

After ranking enriched chunks and selecting the top-$k$ candidates, we construct the final context provided to the LLM through a structured assembly procedure (see Fig.~\ref{fig:pipeline}). This step consists of three stages:
\begin{enumerate}
    \item \textbf{Reference Expansion.}
    For each selected base chunk, we retrieve its reference-closure set (headings and tables obtained via BFS in Phase~2). This ensures that all semantically linked regulatory elements required to instantiate a scenario are included.
    \item \textbf{Intra- and Inter-Chunk De-duplication.}
    Because multiple selected chunks may reference the same paragraph or table, we remove duplicate entries using their stable regulation IDs. This guarantees that each referenced element appears exactly once in the final context and avoids unnecessary token overhead.
    \item \textbf{Canonical Ordering.}
    The resulting set of base chunks and referenced items is sorted according to the regulation's original hierarchical numbering (e.g., 5.2.1 $\prec$ 5.2.1.1 $\prec$ 5.2.2). Tables inherit the position of their defining paragraph ID. This ordering reconstructs the logical flow of the regulation rather than the arbitrary similarity-based retrieval order. The sorted chunks are then concatenated in this canonical order to form the final context window returned to the LLM.
\end{enumerate}
This ordering step is crucial because similarity-based retrieval alone may return relevant fragments in a non-coherent order. By restoring canonical regulation order and merging referenced tables with their defining clauses, RegulaRAG provides the LLM with a logically structured evidence block, reducing hallucinations and improving numerical consistency in generated scenarios.

Algorithm~\ref{alg:context-assembly} formalises the complete context assembly procedure.

\begin{algorithm}[!htbp]
\small
\caption{Reference-Aware Context Assembly (Smart Retrieve \& Rerank)}
\label{alg:context-assembly}

\KwIn{Query $q$; enriched chunk set $\mathcal{C}_E$ where each chunk $c$ carries base text and reference-closure $R(c)$; retrieval encoder $\mathrm{enc}$; retrieval breadth $k$.}

\KwOut{Ordered, de-duplicated context window $W$.}

\BlankLine

$E_q \gets \mathrm{enc}(q)$ \tcp*{embed query}

\ForEach{enriched chunk $c \in \mathcal{C}_E$}{
    $E_c \gets \mathrm{enc}(c.enriched\_text)$\;
    $s_c \gets \mathrm{cosine\_similarity}(E_q,\, E_c)$\;
}

$\mathcal{T} \gets \arg\!\mathrm{top}_k\{\,s_c : c \in \mathcal{C}_E\}$ \tcp*{select top-$k$ chunks}

\BlankLine
\tcp{Step 1 — Reference expansion}
$\mathcal{T}_{\mathrm{exp}} \gets \mathcal{T}$\;
\ForEach{base chunk $c \in \mathcal{T}$}{
    $\mathcal{T}_{\mathrm{exp}} \gets \mathcal{T}_{\mathrm{exp}} \cup R(c)$ 
}

\BlankLine
\tcp{Step 2 — De-duplication }
$\mathcal{T}_{\mathrm{uniq}} \gets \mathrm{deduplicate}(\mathcal{T}_{\mathrm{exp}})$\;

\BlankLine
\tcp{Step 3 — Canonical ordering ; }
$W \gets \mathrm{sort}\!\left(\mathcal{T}_{\mathrm{uniq}},\;\mathrm{key}=\mathrm{regulation\_id}\right)$\;
\Return $W$\;

\end{algorithm}

\subsection{Dataset description}\label{AA}
To evaluate the capabilities of Large Language Models (LLMs) in generating test scenarios from regulatory standards, we constructed a dataset derived from UN Regulation No. 152. This dataset consists of manually extracted and categorized test scenarios relevant to Advanced Emergency Braking Systems (AEBS). The dataset is summarized in table \ref{tab:scenario_summary}. We release the dataset publicly on Hugging Face at \href{https://huggingface.co/datasets/vahidzolf/un_152}{\texttt{vahidzolf/un\_152}}.

\begin{table}[htbp]
\caption{Summary of Test Scenarios by Category and Condition}
\begin{center}
\begin{tabular}{|c|c|c|}
\hline
\textbf{Category} & \textbf{Condition} & \textbf{Num Scenarios} \\
\hline
\multirow{2}{*}{CtoStC} & unladen & 12 \\
                        & laden   & 12 \\
\hline
\multirow{2}{*}{CtoMoC} & unladen & 8  \\
                        & laden   & 7  \\
\hline
\multirow{2}{*}{CtoP}   & MassRun & 10 \\
                        & MaxMass & 10 \\
\hline
\end{tabular}
\label{tab:scenario_summary}
\end{center}
\end{table}

The scenarios are divided into three primary categories, each representing a distinct type of test situation:

\begin{itemize}

\item Car-to-Stationary-Car (CtoStC): These scenarios assess the AEBS functionality when a vehicle approaches a stationary car. The tests include variations in vehicle load (laden/unladen) and different approach speeds (e.g., 20 km/h, 42 km/h, 60 km/h).

\item Car-to-Moving-Car (CtoMoC): This category focuses on test cases where a vehicle interacts with another moving vehicle, considering relative speeds and conditions specified in the standard.

\item Car-to-Pedestrian (CtoP): These scenarios examine the effectiveness of AEBS in detecting and mitigating collisions with pedestrians crossing the road at predefined speeds and trajectories.

\end{itemize}

Each scenario entry in the dataset contains a unique identifier, test title, and a detailed textual description outlining the specific test conditions.

\paragraph{Dataset construction and verification.}
To construct the ground-truth dataset, we systematically reviewed UN Regulation No.\,152 and identified the three scenario categories defined in the standard: CtoStC, CtoMoC, and CtoP.
For each category, we first extracted the general test conditions, which are distributed across multiple sections of the regulation.
For example, the CtoStC conditions include the requirement from Section~5.2.1.4 that ``the subject vehicle shall approach the lead vehicle in a straight line for at least 2\,s before the functional part of the test commences.''
We then combined these general conditions with the scenario-specific numerical requirements for each test case.
Based on this combined information, we created a template scenario for each category, then instantiated each template by varying the relevant parameters according to the values specified in the regulation.
For CtoStC, for instance, we varied approach speed using the values listed in the ``Maximum Relative Impact Speed (km/h) for M1 vehicles'' table and repeated the process separately for laden and unladen vehicle conditions; the corresponding post-condition requirements were filled from the applicable regulation values for the stationary target case.
The same procedure was followed for CtoMoC and CtoP.
After the initial extraction and construction, all resulting scenarios were reviewed for consistency with the regulation text and cross-checked against the original source sections and tables to confirm accuracy.

\subsection{Evaluation system}\label{sec:evalc}

To evaluate the performance of different Large Language Models (LLMs), we employ a structured scoring system based on the dataset we have prepared. The generated test scenarios from each LLM are compared against our manually curated ground truth dataset to assess their accuracy and completeness.

The scoring process begins with the RAG system, which retrieves relevant context from UN Regulation No. 152 to generate test scenarios based on an initial prompt. The output is then directly compared to the manually constructed ground truth scenarios. \\
For evaluation of generated scenarios, we use a different embedding model, namely \textsf{sentence-transformers/all-MiniLM-L6-v2}\footnote{\url{https://huggingface.co/sentence-transformers/all-MiniLM-L6-v2}} from the same sentence-transformers library (v3.4.1), which produces 384-dimensional embeddings.
This model differs from the retrieval model and is computationally lighter, making it well-suited for large-scale pairwise similarity scoring between generated and ground-truth scenarios. We use this model to create vector embeddings for both the generated and ground-truth scenarios. 

The similarity measurement is conducted using cosine similarity. We chose it as it is widely adopted in semantic text comparison tasks due to its efficiency, scale-invariance, and robustness to variations in text length~\cite{reimersSentenceBERTSentenceEmbeddings2019}.

One key challenge in LLM-generated scenarios is the incorrect assignment of numerical values from the regulation text. While LLMs tend to generate the general structure and wording of the test scenarios correctly, they often fail to extract and apply numerical values accurately. For example, a scenario requires the subject vehicle to travel at 20 km/h (with a tolerance of +0/-2 km/h) before braking. This speed value should be derived from the regulation statement: \textit{"Tests shall be conducted with a vehicle travelling at 20, 42, and 60 km/h (with a tolerance of +0/-2 km/h)."} However, LLMs sometimes generate incorrect speed values that are not supported by the regulation text. Similarly, the post-condition requirement, which dictates that the "\textit{AEBS must decrease the vehicle speed to 0 km/h}", should be extracted from the Maximum Relative Impact Speed (km/h) for M1 vehicles table (see Figure \ref{fig:splittable}). LLMs often fail to distinguish between laden and unladen conditions, leading to incorrect post-condition values in the generated scenarios.

However, to ensure a more nuanced evaluation, we incorporate a penalization mechanism that accounts for critical differences between the generated and reference scenarios. This penalization ensures that minor lexical variations are tolerated while significant discrepancies, such as omitted or incorrect details, are appropriately reflected in the final evaluation. A related scoring challenge is that cosine similarity scores tend to be very high, often close to 1.0, due to the fact that the LLM correctly reproduces the scenario template but fails to assign proper numerical values. To address this, we implemented a regex-based matching approach that extracts all numerical values along with the terms laden and unladen from both the generated and ground truth scenarios. The extracted values are then compared using string matching. While cosine similarity might yield a score around 0.95, we introduce a penalty weight $\lambda = 0.2$ applied once per discrepancy identified in the numerical values or laden/unladen conditions. This adjustment ensures that LLMs producing incorrect numerical values receive a final score below 0.9 ($\theta$ in Algorithm~\ref{alg:penalized-scoring} ), which we classify as not equivalent to the ground truth. This refined approach enhances the robustness of our scoring mechanism, ensuring that the most critical numerical details are accurately assessed and penalized when incorrect.

To make scenario matching sensitive to critical compliance attributes that cosine similarity alone often misses (notably numeric parameters and loading conditions), we extract two structured key sets from each scenario text: (i) numeric values and (ii) load/condition terms. Formally, we define two regular-expression families:

 \textbf{Numeric pattern set \(\mathcal{R}_{\#}\).} 
We extract signed integers and decimals (supporting both decimal dots and commas) using:
\[
\mathcal{R}_{\#}:\quad \texttt{(?<!\textbackslash w)[+\textbackslash-]?\textbackslash d+(?:[.,]\textbackslash d+)?}.
\]
Prior to extraction, we normalize the text by (a) converting decimal commas to decimal dots (e.g., \(0,2 \rightarrow 0.2\)), and (b) splitting fused unit strings (e.g., ``2s'' \(\rightarrow\) ``2 s''). 

 \textbf{Load/condition term set \(\mathcal{R}_{\ell}\).} 
We extract regulation-relevant categorical flags as case-insensitive keywords and de-duplicate them:

\begin{align*}
&\hspace*{\fill}\texttt{R}_\ell\texttt{: \textbackslash b(?:unladen|laden|MassRun|MaxMass|}\\
&\hspace*{\fill}\texttt{Maximum\textbackslash s+mass|Mass\textbackslash s+in\textbackslash s+running}\\
&\hspace*{\fill}\texttt{\textbackslash s+order|stationary|moving|stopped)\textbackslash b}
\end{align*}

Given a scenario text \(x\), the extraction function returns
\(K(x) = \big(N(x), L(x)\big)\), where \(N(x)\) is the list of numbers extracted by \(\mathcal{R}_{\#}\) after normalization/masking, and \(L(x)\) is the set of categorical terms extracted by \(\mathcal{R}_{\ell}\).
We then compute a discrepancy score \(\Delta\) between two scenarios by combining:
(i) a tolerance-based one-to-one matching F1 over numeric lists (with absolute tolerance \(\varepsilon_{\mathrm{abs}}\) and relative tolerance \(\varepsilon_{\mathrm{rel}}\)), and 
(ii) Jaccard distance over the categorical term sets. 
This discrepancy is converted into a penalized score. Concretely, for a candidate pair of generated scenario $h$ and ground-truth scenario $g$:
\begin{equation}
  s(h,g) \;=\; s_{\cos}(E_h,\,E_g) \;-\; \lambda\,\Delta\!\bigl(K(g),\,K(h)\bigr)
  \label{eq:penalized-score}
\end{equation}
where $s_{\cos}$ is the cosine similarity between sentence embeddings $E_h$ and $E_g$, $\lambda$ is the penalty weight (default $\lambda=0.2$), and $\Delta$ is the combined numeric/categorical discrepancy defined above. The complete matching and F1 computation procedure is formalised in Algorithm~\ref{alg:penalized-scoring}.

\begin{algorithm}[!htbp]
\small
\caption{Penalized Scoring of Generated Scenarios}
\label{alg:penalized-scoring}

\KwIn{Ground-truth CSV $\mathcal{G}$; Generated CSV $\mathcal{H}$;
threshold $\theta$; penalty weight $\lambda$ (default $\lambda=0.2$); regex sets
$\mathcal{R}_{\#}$ (numbers), $\mathcal{R}_{\ell}$ (load terms).}

\KwOut{Precision, Recall, F1; counts TP/FP/FN.}

\BlankLine

Load all rows from $\mathcal{G}$ and $\mathcal{H}$; keep Unique\_ID and text\;
Initialize $\mathrm{TP} \gets 0$, $\mathrm{FP} \gets 0$, $\mathrm{FN} \gets 0$\;
Mark all $h \in \mathcal{H}$ as unmatched\;

\ForEach{ground-truth scenario $g \in \mathcal{G}$}{
    $E_g \gets \mathrm{encode}(g.text)$\;
    $K_g \gets \mathrm{extract}(g.text;\, \mathcal{R}_{\#} \cup \mathcal{R}_{\ell})$ \tcp*{numbers + load flags}
    
    $s^\star \gets -\infty$, $h^\star \gets \varnothing$\;
    
    \ForEach{generated scenario $h \in \mathcal{H}$}{
        $E_h \gets \mathrm{encode}(h.text)$\;
        $K_h \gets \mathrm{extract}(h.text;\, \mathcal{R}_{\#} \cup \mathcal{R}_{\ell})$\;
        
        $s_{\cos} \gets \mathrm{cosine\_similarity}(E_h, E_g)$\;
        $\Delta \gets \mathrm{diff}(K_g, K_h)$ \tcp*{count discrepancies}
        
        $s \gets s_{\cos} - \lambda\,\Delta$ \tcp*{apply penalty: $\lambda$ per discrepancy}
        
        \If{$s > s^\star$}{
            $s^\star \gets s$\;
            $h^\star \gets h$\;
        }
    }
    
    \If{$s^\star \ge \theta$}{
        Mark $h^\star$ as matched\;
        $\mathrm{TP} \gets \mathrm{TP} + 1$\;
    }
    \Else{
        $\mathrm{FN} \gets \mathrm{FN} + 1$\;
    }
}

$\mathrm{FP} \gets |\{h \in \mathcal{H} : h \text{ matched}\}| - \mathrm{TP}$\;

$\mathrm{Precision} \gets \frac{\mathrm{TP}}{\mathrm{TP} + \mathrm{FP}}$\;
$\mathrm{Recall}    \gets \frac{\mathrm{TP}}{\mathrm{TP} + \mathrm{FN}}$\;
$\mathrm{F1} \gets \frac{2 \cdot \mathrm{Precision} \cdot \mathrm{Recall}}
                        {\mathrm{Precision} + \mathrm{Recall}}$\;

\end{algorithm}

\paragraph{Metric generalizability.}
The evaluation framework comprises two layers with different degrees of domain specificity.
The penalized F1 metric (Eq.~\ref{eq:penalized-score}) is structurally regulation-agnostic: the cosine-similarity backbone, tolerance-based numeric matching, and Jaccard-distance categorical comparison can in principle be applied to any domain that produces structured textual artefacts with verifiable numerical parameters (e.g., pharmaceutical dosage specifications, financial compliance documents, or aviation checklists).
The domain-specific elements are confined to the two regex families: $\mathcal{R}_{\#}$ captures generic numeric expressions and requires minimal adaptation across domains, whereas $\mathcal{R}_{\ell}$ encodes UN-152-specific load-condition flags (\textit{laden}, \textit{unladen}, \textit{MassRun}, etc.) that must be redefined to match the categorical vocabulary of another standard.
The penalty weight $\lambda$ is a tunable hyperparameter, not a regulation-specific constant.
Similarly, the Meta-Score ($\bar{F}_1 - \sigma$) is fully domain-agnostic: it summarises mean performance and cross-condition stability using any compatible base metric and is applicable whenever a system is evaluated across multiple structured task categories.
In summary, adapting the metric to a new regulatory domain requires (i) replacing $\mathcal{R}_{\ell}$ with domain-relevant categorical keywords, and (ii) re-tuning $\lambda$ and the matching threshold $\theta$ on a representative sample; the rest of the framework transfers without modification.

\paragraph{Compliance-oriented metrics.}
Automotive regulations do not define a universal compliance-accuracy metric; compliance is usually assessed through regulation-specific pass/fail criteria, tolerances, and traceability. In our evaluation, these aspects are encoded in the manually constructed ground-truth scenarios from UN Regulation No.~152. Thus, the reported F1-score serves as a proxy for scenario-level compliance, while the penalized scoring explicitly checks critical numerical values (\(\mathcal{R}_{\#}\)) and loading conditions (\(\mathcal{R}_{\ell}\)). Dedicated metrics such as Requirement Coverage, Parameter Accuracy, and Traceability Coverage would provide additional insight and are left as future work, informed by recent advances in LLM-based requirements traceability~\cite{alturayeifTraceLLMLeveraging2026,etezadiClassifierPromptCase2026,folorunshoAIDrivenTestCase2026}.

\section{Experimental Setup}\label{exp_setup}

\subsection{Language Models}
We report results with three representative LLMs:
(i) \textbf{gpt-4o-2024-11-20} accessed via the OpenAI API\footnote{\url{https://platform.openai.com/docs/api-reference}};
(ii) \textbf{Llama\,3.3\,70B (``llama-3.3-70b-versatile'')} served through the Groq API\footnote{\url{https://console.groq.com/docs/models}};
and (iii) \textbf{DeepSeek-chat} (API model identifier: \texttt{deepseek-chat}) invoked via the vendor’s direct API\footnote{\url{https://api-docs.deepseek.com/}}.
RegulaRAG operates upstream of generation: it determines which parts of the regulation reach the prompt, and the language model is an interchangeable component of the pipeline rather than part of the contribution.
The three models were therefore selected to span distinct deployment settings --- a proprietary API model, an openly available model, and a low-cost API model --- rather than to identify a best-performing generator, and pinned model snapshots (e.g., \texttt{gpt-4o-2024-11-20}) were preferred so that runs remain reproducible.
Because every RAG system is evaluated under the same generator with identical prompts, scenario definitions, and decoding settings (Section~\ref{sec:rag_comparison}), the comparison isolates the retrieval strategy, and a different or newer generator can be substituted without any change to the pipeline.
DeepSeek offers both a \emph{reasoning} model (DeepSeek-reasoner) and a \emph{chat} model;
We adopt the chat variant because, in our setting, it achieves competitive accuracy while being over 12$\times$ cheaper than the reasoning model (\$0.14
  vs.\ \$1.74 per million input tokens on a cache miss), making it the practical choice for large-scale pipeline execution.
Note that the DeepSeek API does not expose a pinned checkpoint version for the \texttt{deepseek-chat} endpoint; the underlying model may be updated by the provider without a version change in the model string.
All models are run with low temperature (\(\tau=0.1\)) to reduce output variance; decoding settings are held fixed
across methods.
Retrieval hyperparameters (\(top\_k\) and \(chunk\_size\)) are shared across all LLMs and determined via grid search, as described in Section~\ref{sec:gridSearch}.

\subsection{Evaluation Protocol}
We report Precision, Recall, and F1 at the scenario level, using F1
as the primary metric. Each failing run is automatically retried up to two additional times and marked \texttt{ok} upon the first successful completion. For corpus-scaling experiments (Section~\ref{sec:scalability}), we additionally record end-to-end runtime.
Scenario matching uses a cosine similarity threshold of $\theta = 0.9$: a generated scenario is counted as a true positive only if its penalized similarity to the best-matching ground-truth scenario meets or exceeds this value. Numeric values within scenarios are compared using an absolute tolerance $\varepsilon_{\mathrm{abs}} = 0.05$ and a relative tolerance $\varepsilon_{\mathrm{rel}} = 0.05$ (5\%).
Due to resource constraints, each (RAG system, LLM, scenario-type) configuration was evaluated in a single run; trial-level dispersion statistics across repeated executions are therefore not reported. This limitation is explicitly acknowledged in Section~\ref{sec:conclusion}.

\subsection{Compute and Reproducibility}
Experiments were executed on a Linux workstation with an NVIDIA GeForce GTX~1080~Ti
(11\,GB GDDR5X), an Intel\textsuperscript{\textregistered} Core\texttrademark{} i7-3770 CPU @ 3.40\,GHz,
and 32\,GB RAM. The software stack uses Python~3.10.12. Random seeds (where applicable),
prompt templates, and decoding parameters are fixed across conditions.
The ground-truth scenario dataset is publicly available on Hugging Face (\href{https://huggingface.co/datasets/vahidzolf/un_152}{\texttt{vahidzolf/un\_152}}; see Section~\ref{AA}).
Code, prompt templates, and evaluation CSVs necessary to reproduce the results are publicly available in the project repository\footnote{\url{https://gitlab.lrz.de/vahidev/retrieval-augmented-generation}}; API keys are required for commercial models.

Two distinct sentence-transformer models from the \textsf{sentence-transformers} library (v3.4.1) are used at separate pipeline stages:
\begin{itemize}
  \item \textbf{Retrieval:} \textsf{all-mpnet-base-v2} (768-dim) — embeds enriched chunks and queries for similarity-based top-$k$ selection in Phase~3.
  \item \textbf{Evaluation:} \textsf{all-MiniLM-L6-v2} (384-dim) — computes pairwise cosine similarity between generated and ground-truth scenarios in the penalized scoring metric.
\end{itemize}

\subsection{Prompting System}
\label{subsec:prompting}

To obtain consistent outputs from all language models (LLMs), we used identical prompts across all models without applying any advanced prompting techniques. A single system prompt was used to describe the general task, and three user prompts were designed, each targeting a specific type of scenario: Car-to-Stationary-Car, Car-to-Moving-Car, and Car-to-Pedestrian.

These user prompts requested the generation of scenarios according to the UN Regulation No. 152, each including a single example to act as a structural template. You can find the complete user prompts in Appendix \ref{appendix:user_prompts}. We deliberately avoided including instructions on how to derive the scenarios in prompts, to keep the LLMs task-focused and to support automated evaluation against a ground truth dataset. The example format ensured that all outputs followed a consistent structure, enabling effective comparison and scoring.

\begin{figure}[t]
\centering
\begin{minipage}{0.95\linewidth}
\begin{quotation}\label{lst:system_prompt}
\textit{\textbf{User Prompt for Car-to-Moving-Car (CtoMoC)}}\\
\textit{You are an autonomous driving engineer specializing in AEBS testing for M1 category vehicles. You must create test scenarios following UN Regulation No. 152 exactly as specified, with particular attention to:
- Vehicle positioning and approach conditions
- Precise speed requirements including tolerances
- Time To Collision (TTC) specifications
- Loading conditions (laden/unladen)
- Clear post-condition requirements}
\end{quotation}
\end{minipage}
\caption{System prompt for Car-to-Moving-Car (CtoMoC)}
\label{fig:system-prompt-ctomoc}
\end{figure}

\section{Results and Experiments}\label{res_and_exp}
\subsection{Experiment Design}\label{exp_design}

To thoroughly evaluate our system, a staged experimental design was adapted. The search space is large because multiple pipeline components interact; therefore, we decompose the problem into controllable factors and explore them systematically rather than exhaustively. Accordingly, the following evaluation dimensions was defined, each capturing a distinct aspect of the RAG pipeline that must be scrutinized to understand performance, robustness, and scalability:

\begin{enumerate}
  \item \textbf{RAG system}
      We evaluated six RAG systems: \textbf{RegulaRAG} (our method), \textbf{R\&R+RCS}, \textbf{NoRAG}, \textbf{OpenAI-RAG}, \textbf{HippoRAG}, and \textbf{Hybrid}.

  \item \textbf{Scenario family} \quad
  Car-to-Stationary-Car (CtoStC), Car-to-Moving-Car (CtoMoC), and Car-to-Pedestrian (CtoP). Section \ref{sec:rag_comparison} reports the corresponding results of this and previous item. 

  \item \textbf{Retrieval breadth} (\(top\_k\)) \quad
  \(top\_k \in \{10,\,15,\,20,\,25,\,30\}\).

  \item \textbf{Chunking granularity} (\(chunk\_size\)) \\ To assess the impact of chunk granularity on RAG performance, the chunk size  varied across a range of values. Section \ref{sec:gridSearch} reports the corresponding results.

  \item \textbf{Source size (corpus scale)} \quad
  To assess robustness under semantically similar distractors, we progressively expand the retrieval corpus using eight UN regulations in the automotive software domain, including UN Regulation No. 152. This setup enables a systematic evaluation of the scalability, strengths, and weaknesses of our method relative to the strongest competing RAG baseline. section \ref{sec:scalability} discusses the result of this experiment. 
  
  \item \textbf{Language model (LLM)} \quad
  GPT-4o was used as a widely adopted commercial model, and include DeepSeek-Chat and Llama~3.3 to probe cost/latency and open-weight generalization.
\end{enumerate}

An exhaustive grid over all factors would be computationally prohibitive and confounds effects across dimensions. Instead, we adopt a progressive design that (i) first identifies stable hyperparameters for retrieval by performing grid search, (ii) then compares RAG systems under matched conditions (to isolate retrieval effects), and (iii) finally stresses the systems with larger corpora (to assess scaling and distractor robustness). This ordering controls nuisance variability and yields conclusions that are both fair and reproducible. Following is the sequence of experiments:

\subsubsection{Hyperparameter selection via grid search}\label{sec:gridSearch}
\begin{figure*}[t!]
\centering
\includegraphics[width=0.6\textwidth]{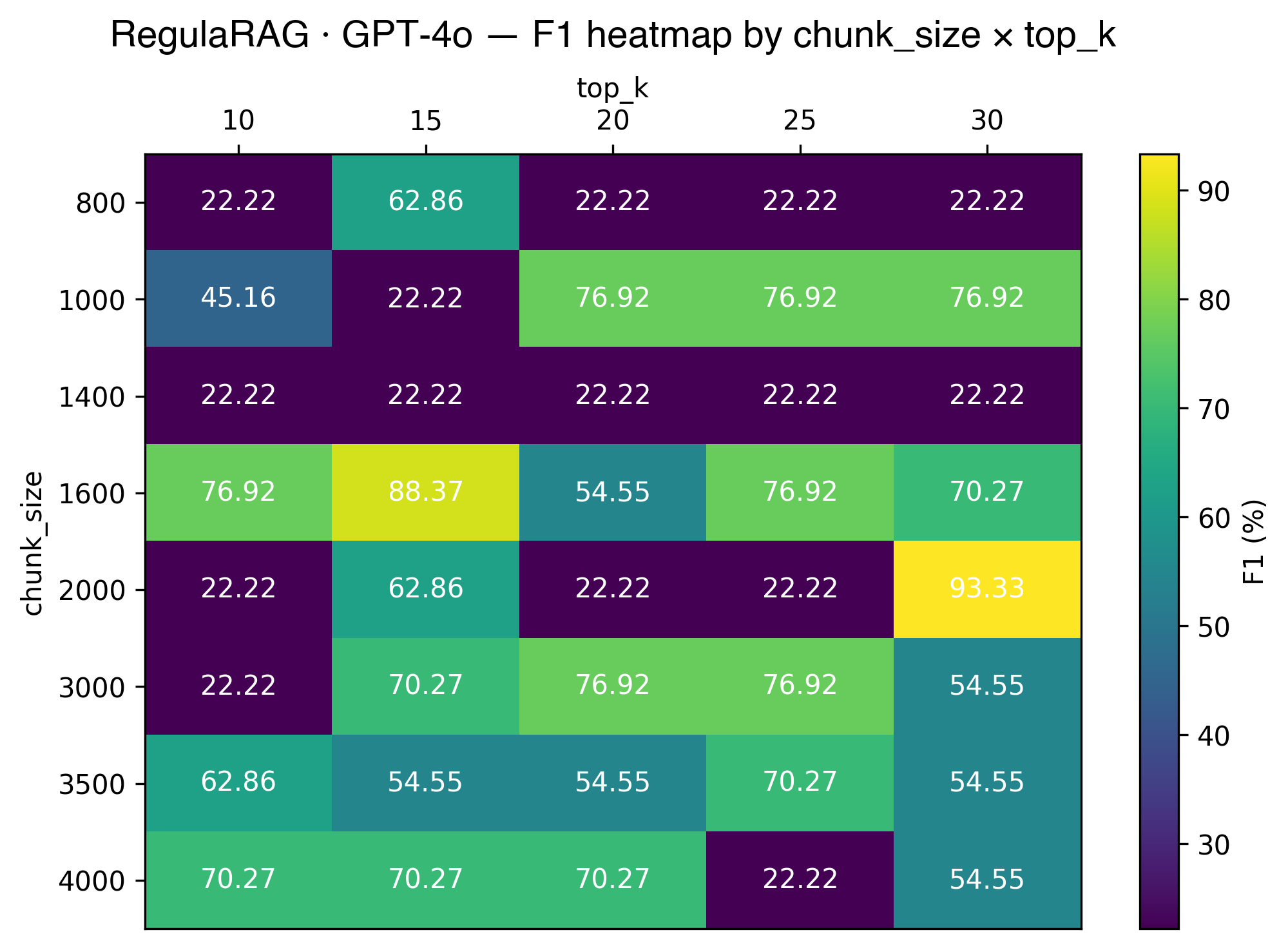}
\caption{\textbf{Hyperparameter Selection}\label{fig:top_k_chunk_size}}
\end{figure*}

We estimate robust retrieval hyperparameters by performing a full grid sweep on the CtoStC scenario using UN-152 as the sole source. We evaluate:\\
\(top\_k \in \{10,15,20,25,30\}\) \\ 
\(chunk\_size \in \{800,1000,1400,1600,2000,3000,4000\}\)
for our RegulaRAG method when using GPT-4o for the final scenario generation. Figure \ref{fig:top_k_chunk_size} presents F1 as a heatmap over the \((chunk\_size, top\_k)\) grid.
This pattern is consistent with RAGGED~\cite{240309040RAGGEDTowards}, which finds that readers exhibiting an improve-then-plateau response to retrieval depth --- a trend GPT-4o follows alongside GPT-3.5-turbo --- incur substantially smaller performance loss when operating away from their individually optimal $top\_k$ than noise-sensitive readers (e.g., LLaMA, Claude); this may partly explain why $top\_k{=}30$ remained effective across the eight-regulation scalability experiment (Section~\ref{sec:scalability}).
The resulting response surface shows broad plateaus with a few sharp maxima. This plot reveals that, despite a few outliers in the heatmap, the overall performance trend stabilizes around \((top\_k{=}30,\;chunk\_size{=}2000)\).  
At this configuration the F1-score reaches its peak before gradually degrading for larger values, indicating that it represents a near-optimal balance between retrieval breadth and chunk granularity.
To ensure a fair comparison, all retrieval-based baselines (R\&R+RCS, OpenAI-RAG, HippoRAG, and Hybrid) were subsequently evaluated using the same \(top\_k{=}30\) value identified here.

\subsubsection{RAG system comparison.}\label{sec:rag_comparison}
  With \((top\_k^{\ast},\,chunk\_size^{\ast})\) fixed,  we evaluated six Retrieval-Augmented Generation (RAG) pipelines across all three scenario families (CtoStC, CtoMoC, and CtoP) on UN--152:
\begin{itemize}
  \item \textbf{RegulaRAG} --- our method: SmartChunking with reference-aware enrichment + smart retrieval:  Smart Retrieve and Rerank which
is integrated into our RegulaRAG pipeline. It compares the enriched chunks (including referenced content) to the query, then gathers the corresponding original chunks and appends the reference chunks after deduplication. 
  \item \textbf{R\&R+RCS (rr\_rcs)} --- baseline: RecursiveCharacterSplitter (RCS) + simple retrieval: this method uses the \textsf{all-mpnet-base-v2}\footnote{\url{https://huggingface.co/sentence-transformers/all-mpnet-base-v2}} model, the same model as the one we used for our regulaRAG method, to embed the chunks and queries.  The standard version performs retrieval directly over the original chunks and returns the top-k chunks based on similarity.
  \item \textbf{OpenAI-RAG (openai\_rag)} --- baseline: RCS + OpenAI embeddings + Chroma: In this method, we use OpenAI’s \\
  \textsf{text-embedding-ada-002} model\footnote{\url{https://platform.openai.com/docs/guides/embeddings}} to generate chunk embeddings, which are stored in a Chroma vector database. Retrieval is performed using LangChain’s vector store-backed retriever with the \\similarity\_score\_threshold search type and parameters \textsf{search\_kwargs=\{"k": 30, "score\_threshold": 0.1\}}.

  \item \textbf{HippoRAG} --- baseline: As a state-of-the-art knowledge graph-based RAG method, HippoRAG \cite{hipporag} is included to benchmark our retrieval pipeline against recent graph-based systems. It has shown competitive performance in recent benchmarks.

  \item \textbf{NoRAG} --- reference baseline: Direct LLM generation without any retrieval context. The model receives only the user prompt and scenario-format example, with no regulatory document chunks provided. Included to measure the scenario-generation capability embedded in model weights alone, quantifying the marginal benefit of the retrieval pipeline.

  \item \textbf{Hybrid} --- baseline: A hybrid retrieval pipeline that combines sparse keyword-based (BM25) retrieval with dense semantic (embedding-based) retrieval over RCS-generated chunks, merging the two ranked lists before context assembly.
\end{itemize}

This design isolates the contribution of the chunking and retrieval strategy by keeping the scenario definitions, prompts, and model-specific decoding settings constant. We report F1 score as the primary metric and track precision/recall for diagnostic completeness. The goal is to establish whether RegulaRAG’s reference-aware enrichment translates into consistent gains across distinct task conditions.
The results are presented in Table \ref{tab:meta_scores}. $\uparrow$ arrows are for metrics where higher is better and $\downarrow$ arrows are for metrics where lower is better.\\ 
To capture both the average performance and the stability of each RAG system, 
we introduce a \emph{Meta-Score}, computed from the F1-scores of the three evaluated 
LLMs (GPT-4o, DeepSeek-chat, and LLaMA-3.3). While Algorithm~\ref{alg:penalized-scoring} 
described the penalized evaluation of individual scenarios which yields the F1-score, our goal here is to obtain a metric that also reflects the variance of LLM performance and summarizes how consistently a given RAG--LLM pipeline behaves across different scenario categories. The Meta-Score therefore provides a more holistic indicator of system-level robustness.

Let $\bar{F}_1$ denote the mean of the three F1 values for 3 types of scenarios for a given RAG system,
and let $\sigma$ denote their standard deviation. 
The Meta-Score is then defined as:
\[
\text{MetaScore} = \bar{F}_1  - \sigma.
\]
We propose the Meta-Score as a robustness-adjusted aggregation designed for this study; it is not a standard benchmark metric in RAG evaluation. Its formulation is inspired by the classical mean–variance tradeoff principle~\cite{markowitz1952portfolio} and the one-standard-error model-selection heuristic of~\cite{hastie2009elements}, both of which penalize high-variance estimators by combining mean performance with a dispersion penalty. We employ the coefficient of variation (CV) as a normalized measure of relative variability, a well-established statistical tool for comparing dispersion across scales.

This formulation rewards RAG systems with high average F1 performance while penalizing those with high variability across the three LLMs, thereby reflecting both accuracy and robustness. Negative values in Table~\ref{tab:meta_scores} arise naturally from the definition of the Meta-Score.When a RAG system produces highly inconsistent results across the three scenario categories like scoring near~0 on one task while performing moderately on others, the variance term becomes large and dominates the average performance, yielding a negative Meta-Score.  

Table~\ref{tab:meta_scores} reveals three clear patterns.  
First, \textbf{RegulaRAG} consistently achieves the strongest overall performance and stability across all language models.  
Across GPT-4o, DeepSeek-chat, and LLaMA-3.3, RegulaRAG attains the highest average Meta-Score of \textbf{82.99}, clearly outperforming all competing methods. In comparison, the second-best system (NoRAG) achieves an average Meta-Score of 57.94, followed by Hybrid (56.71) and HippoRAG (55.75).  

This corresponds to an improvement of approximately \textbf{43\%}.
At the individual model level, RegulaRAG maintains consistently high Meta-Scores across all LLMs (80.99 for GPT-4o, 87.90 for DeepSeek-chat, and 80.08 for LLaMA-3.3), whereas competing methods exhibit substantial variability and, in some cases, severe degradation (e.g., negative Meta-Score for DeepSeek-chat under R\&R+RCS).  

\begin{figure*}[t!]
\centering
\includegraphics[width=0.6\textwidth]{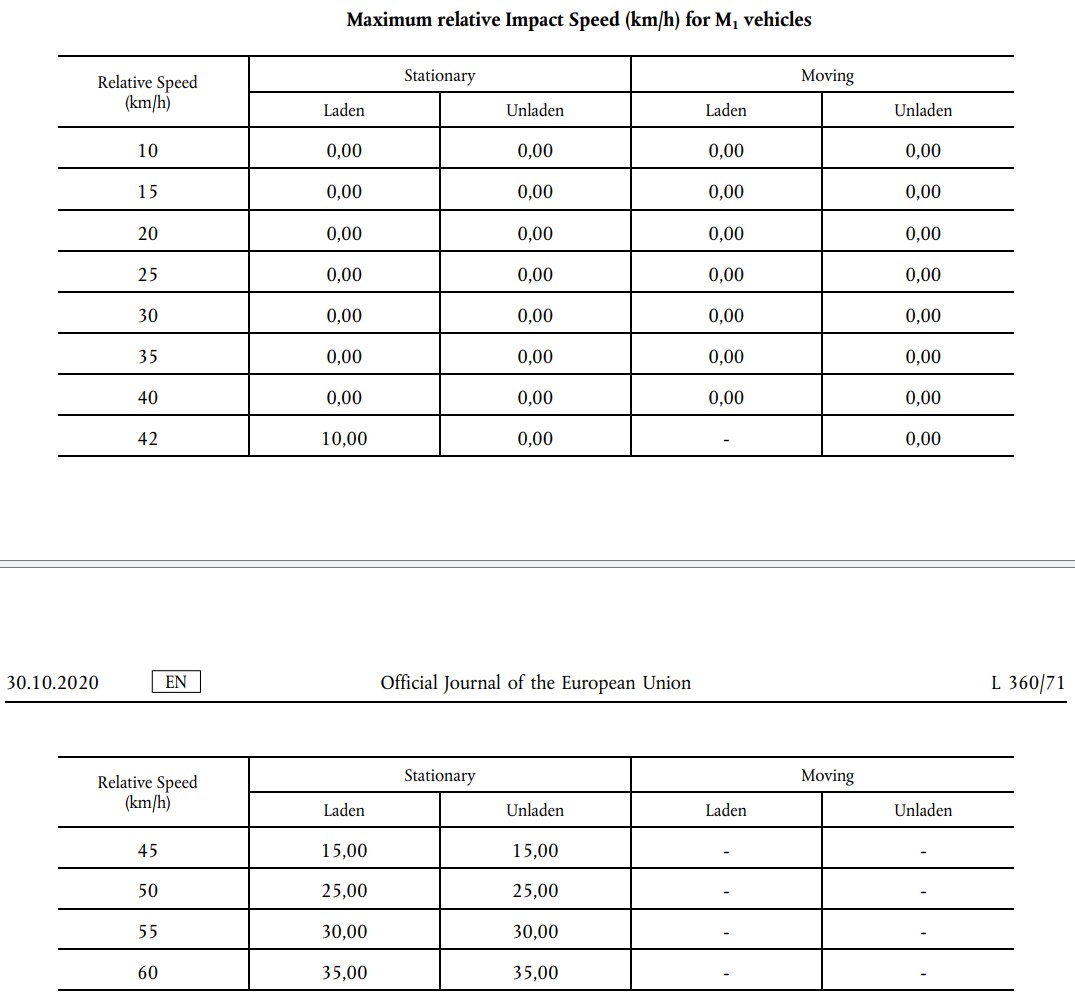}
\caption{\textbf{Example of split table}\label{fig:splittable}}
\end{figure*}

\begin{figure*}[t!]
\centering
\includegraphics[width=0.6\textwidth]{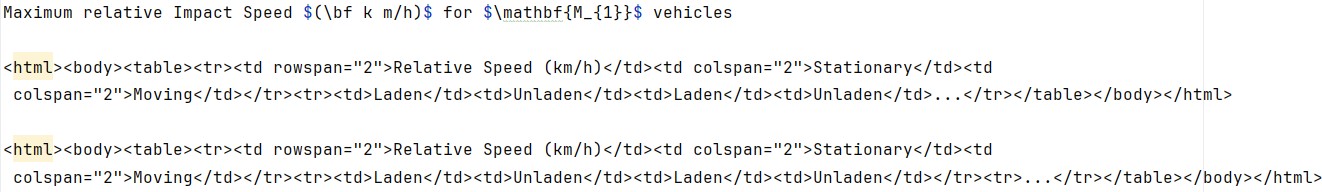}
\caption{\textbf{Converted table (HTML) after PDF split}\label{fig:html_table}}
\end{figure*}

The baseline comparison also provides partial ablation evidence for RegulaRAG’s core components . R\&R+RCS shares the same embedding model (\textsf{all-mpnet-base-v2}) and the same RecursiveCharacterSplitter chunking as RegulaRAG but omits reference-aware BFS enrichment and smart reranking; the large Meta-Score gap between RegulaRAG and R\&R+RCS (82.99 vs.\ 8.88, averaged across LLMs) therefore isolates the joint contribution of those two components. NoRAG, which supplies the LLM with no retrieval context at all, establishes a lower bound on scenario-generation capability embedded in model weights alone (average Meta-Score 57.94); comparing it with RegulaRAG quantifies the marginal value of the entire retrieval pipeline. A fully factorial component-level ablation—isolating BFS reference-closure depth, deduplication, and canonical reordering individually—remains as future work.\\
A manual inspection of retrieved contexts shows that the presence and correct ordering of the necessary table chunks is the primary determinant of high F1. Simply retrieving a relevant table slice is insufficient when the original PDF table is split across pages (Fig.~\ref{fig:splittable}) and subsequently appears as multiple HTML tables (Fig.~\ref{fig:html_table}).
To address this, RegulaRAG uses a rule-based heuristic for table detection and reconstruction. Specifically, when a paragraph contains cues such as ``the following table,'' we treat this as an explicit pointer to subsequent table content. In the Markdown/HTML output produced by Mineru, tables are marked by \texttt{<table>} tags. We therefore collect the consecutive \texttt{<table>} blocks that immediately follow the referring paragraph and attach them to that paragraph as referenced table metadata. This heuristic is particularly useful for PDF tables that are sliced across pages and converted into multiple adjacent HTML table fragments: by grouping consecutive table tags and preserving their original order, the system reconstructs the logical table content before retrieval and prompt assembly.\\
RegulaRAG’s reference-aware chunk reordering ensures these fragments are placed adjacently and in the correct logical order, improving recall and reducing hallucinated numerical substitutions. 

Second, the three scenario categories differ markedly in difficulty. \emph{CtoP} remains the easiest because it is anchored in a self-contained regulatory section (section 5.2.2. Car to pedestrian scenario in \cite{unionRegulationNo1522020} ) with a dedicated table.  In contrast, \emph{CtoStC} and \emph{CtoMoC} require integrating information dispersed 
across multiple paragraphs and shared tables, making them more sensitive to retrieval errors.

Third, the language models respond differently to retrieval conditions.  
DeepSeek-chat performs competitively under RegulaRAG but becomes highly unstable 
under weaker retrieval methods, as reflected by large variance and several negative Meta-Scores for R\&R+RCS and OpenAI-RAG.  
LLaMA-3.3 shows consistent improvement under RegulaRAG but struggles substantially 
with baselines, especially in scenarios requiring precise table interpretation and 
paragraph-level aggregation.  
GPT-4o displays the most stable behavior under RegulaRAG, yet—even for GPT-4o—the 
baseline RAG systems exhibit large variability across tasks.

Another reason that we got better results comparing to the baseline methods is that, thanks to our mechanism for untangling the chain of references distributed across different chunks, the input query is then compared against the entire chain of referenced content along with the chunk's original text, thereby increasing the likelihood of selecting the most relevant chunks. For example, when the input query pertains to a Car-to-Pedestrian scenario in UN Regulation No. 152, a standard chunking and retrieval method may select paragraph 5.2.2, but miss the ``Maximum Impact Speed (km/h) for M1 vehicles'' table, which contains the required speed values. This happens because the table shares no explicit keywords or semantic similarity with the query. In contrast, our method detects that paragraph 5.2.2 contains phrases like ``the following table,'' prompting a search for and attachment of that table to the chunk. This increases the likelihood of retrieving the chunk and provides the LLM with all necessary data to generate complete scenario variants.

\begin{table*}[t]
\centering
\caption{Performance, Stability, and Meta-Score Across RAG Systems and Models}
\label{tab:meta_scores}
\renewcommand{\arraystretch}{1.15}
\setlength{\tabcolsep}{4pt}

\begin{tabular}{|c|c|
                c|c|c|
                c|c|
                c|c|}
\hline
\textbf{RAG System} & \textbf{Model} &
\textbf{CtoStC~($\uparrow$)} &
\textbf{CtoMoC~($\uparrow$)} &
\textbf{CtoP~($\uparrow$)} &
\textbf{Mean~($\uparrow$)} &
\textbf{stdev~($\downarrow$)} &
\textbf{Meta-Score~($\uparrow$)} &
\textbf{Avg.\ 3 LLMs~($\uparrow$)} \\
\hline \Xhline{1pt}

\multirow{3}{*}{RegulaRAG}
 & GPT-4o        & 76.92 & 96.55 & 100.00 & 91.16 & 10.16 & 80.99 & \textbf{82.99} \\
 & DeepSeek-chat & 85.71 & 96.55 & 97.44  & 93.23 &  5.33 & 87.90 &       \\
 & Llama-3.3     & 76.92 & 96.55 & 91.89  & 88.45 &  8.37 & 80.08 &       \\

\hline \Xhline{1pt}

\multirow{3}{*}{R\&R+RCS}
 & GPT-4o        &  8.00 & 88.89 & 91.89  & 62.93 & 38.86 & 24.07  &  8.88 \\
 & DeepSeek-chat &  0.00 &  0.00 & 97.44  & 32.48 & 45.93 & -13.45 &       \\
 & Llama-3.3     & 76.92 &  0.00 & 91.89  & 56.27 & 40.26 & 16.01  &       \\

\hline \Xhline{1pt}

\multirow{3}{*}{NoRAG}
 & GPT-4o        & 22.22 & 96.55 & 100.00 & 72.92 & 35.88 & 37.04 & 57.94 \\
 & DeepSeek-chat & 73.68 & 96.55 & 97.44  & 89.22 & 11.00 & 78.23 &       \\
 & Llama-3.3     & 50.00 & 96.55 & 91.89  & 79.48 & 20.93 & 58.55 &       \\

\hline \Xhline{1pt}

\multirow{3}{*}{HippoRAG}
 & GPT-4o        & 62.86 & 96.55 & 97.44  & 85.62 & 16.10 & 69.52 & 55.75 \\
 & DeepSeek-chat & 90.91 & 80.00 & 97.44  & 89.45 &  7.19 & 82.26 &       \\
 & Llama-3.3     & 73.68 &  0.00 & 91.89  & 55.19 & 39.73 & 15.46 &       \\

\hline \Xhline{1pt}

\multirow{3}{*}{OpenAI-RAG}
 & GPT-4o        & 62.86 & 94.00 & 99.15  & 85.33 & 16.03 & 69.30 & 43.53 \\
 & DeepSeek-chat & 30.30 &  0.00 & 97.44  & 42.58 & 40.72 &  1.86 &       \\
 & Llama-3.3     & 75.79 & 55.37 & 91.89  & 74.35 & 14.94 & 59.41 &       \\

\hline \Xhline{1pt}

\multirow{3}{*}{Hybrid}
 & GPT-4o        & 54.55 & 96.55 & 97.44  & 82.85 & 20.01 & 62.83 & 56.71 \\
 & DeepSeek-chat & 88.37 & 96.55 & 97.44  & 94.12 &  4.08 & 90.04 &       \\
 & Llama-3.3     & 85.71 &  0.00 & 91.89  & 59.20 & 41.94 & 17.26 &       \\

\hline \Xhline{1pt}
\end{tabular}
\end{table*}

\paragraph{Qualitative example.}
To illustrate the end-to-end pipeline, consider a CtoStC scenario from UN Regulation No.~152.
Figure~\ref{fig:sample_chain} shows how the relevant conditions are distributed across cross-linked sections: the test speed values reside in a table, and the approach constraint is stated in other Section.
Figure~\ref{fig:splittable} shows how the speed table is split across two PDF pages and emerges as separate HTML fragments after conversion.
RegulaRAG's SmartChunking detects the ``the following table'' cue, attaches the consecutive HTML table fragments, and add it as metadata to the original chunk. The Smart Retrieve and Rerank module then retrieves the enriched chunk,and with this assembled context, the LLM generates the scenario.
Table~\ref{tab:qualitative_example} shows the expected output (ground-truth entry
\texttt{Test\_\allowbreak CtoStC\_\allowbreak unladen\_10} from the published dataset, Section~\ref{AA}).
Without reference-aware enrichment the speed table is not retrieved, and the LLM either omits the speed value or hallucinates one which is the primary failure mode described in Section~\ref{sec:failure}.

\begin{table}[H]
\centering
\caption{Ground-truth scenario \texttt{Test\_\allowbreak CtoStC\_\allowbreak unladen\_10} — expected LLM output for a CtoStC test at 10~km/h (unladen).}
\label{tab:qualitative_example}
\small
\begin{tabular}{|p{0.88\linewidth}|}
\hline
Two vehicles of Category M1 AA saloon shall be positioned. The subject vehicle that performs the braking and the lead vehicle. \\
\hline
Both vehicles should face in the same direction of travel. \\
\hline
The subject vehicle shall approach the lead vehicle in a straight line for at least 2~s before the functional part of the test commences. \\
\hline
The subject vehicle should travel at the speed of 10~km/h (with a tolerance of $+$0/$-$2~km/h) when the vehicle brakes. \\
\hline
The functional part of the test shall start at a distance corresponding to a Time To Collision (TTC) of at least 4~seconds from the target. \\
\hline
The subject vehicle is unladen. \\
\hline
The lead vehicle is stationary. \\
\hline
Post-condition requirements: \\
\hline
When the system is activated, the AEBS shall decrease the speed to 0~km/h. \\
\hline
\end{tabular}
\end{table}

\subsection{input document scalability}\label{sec:scalability}

To assess robustness under growing evidence, we gradually expand the retrieval corpus from a single regulation (152) to bundles containing eight documents. The evaluation pool contains eight UN regulations of varying length and structural complexity:
\begin{itemize}
  \item No.\ 10: Electromagnetic compatibility \cite{EURLex42012X092001EURLex}
  \item No.\ 79: Steering equipment \cite{EURLex42008X052701EURLex}
  \item No.\ 130: Lane Departure Warning System (LDWS) \cite{EURLex42014X061801EURLex}
  \item No.\ 140: Electronic Stability Control (ESC) \cite{EURLex42018X1592EURLex}
  \item No.\ 152: AEBS (primary target corpus) \cite{EURLex42020X1597EURLex}
  \item No.\ 155: Cybersecurity and CSMS \cite{EURLex42021X0387EURLex}
  \item No.\ 156: Software update and SUMS \cite{EURLex42021X0388EURLex}
  \item No.\ 13-H: Braking (passenger cars) \cite{EURLex42023X0401EURLex}
\end{itemize}
 We fix
\(LLM^{\ast}=\text{GPT-4o}\), \(top_k^{\ast}=30\), and \(chunk\_size^{\ast}=3000\), varying only
the corpus size. This experiment isolates the effects of semantic clutter and cross-document
interactions on retrieval stability.

Alongside \textsc{RegulaRAG}, we evaluate \textsc{HippoRAG}, a knowledge-graph-driven system that performs LLM-based NER, triple extraction, and graph-structured retrieval. Comparing them under identical scaling conditions highlights contrasting computational behaviors. We include HippoRAG because it trails RegulaRAG across all categories in Table \ref{tab:meta_scores}. 

\begin{figure}[H]
\centering
\includegraphics[width=0.9\linewidth]{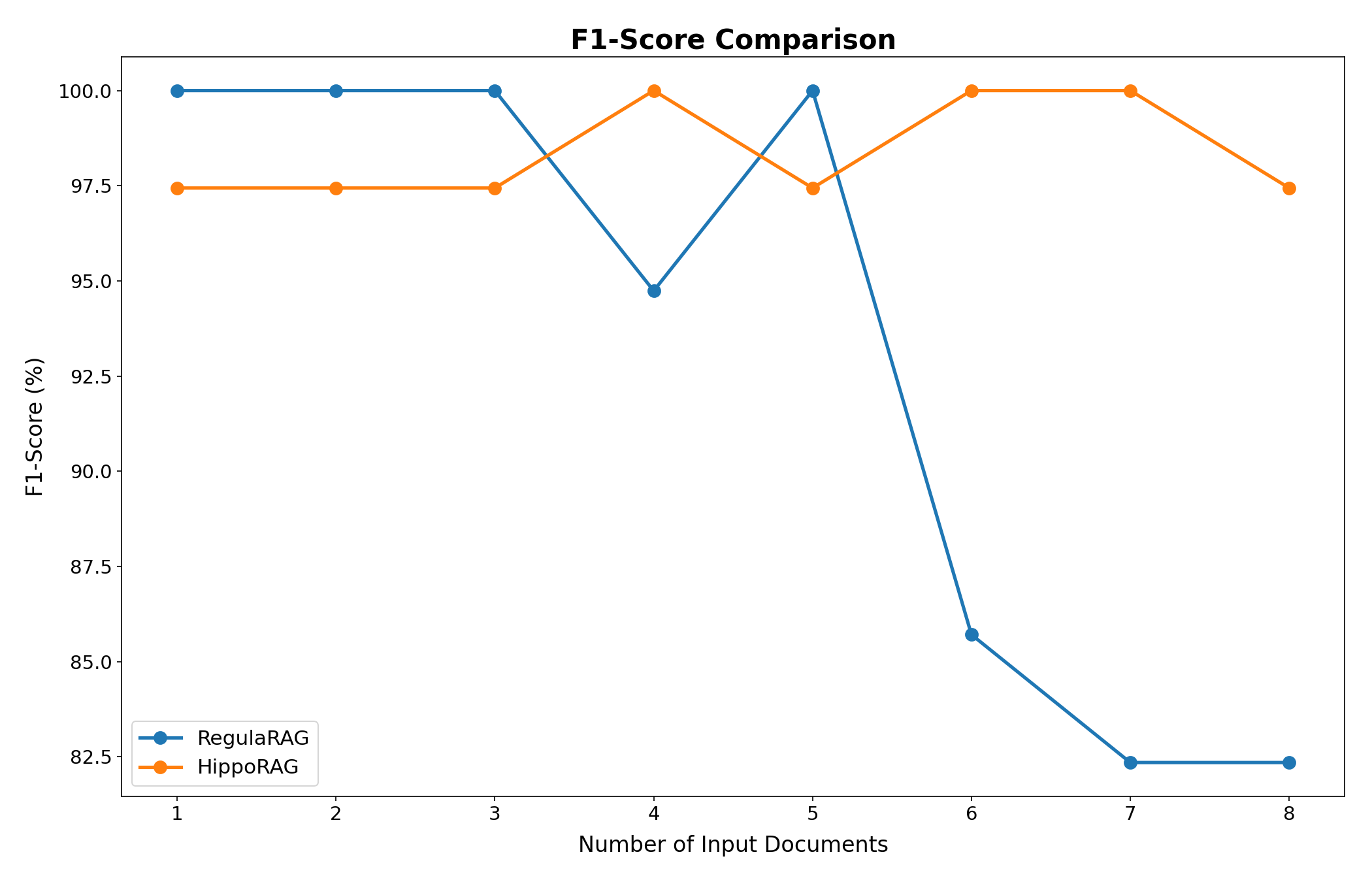}
\caption{F1-score scaling of \textsc{RegulaRAG} and \textsc{HippoRAG} as corpus size increases.}
\label{fig:scaling_f1}
\end{figure}

\begin{figure}[H]
\centering
\includegraphics[width=0.9\linewidth]{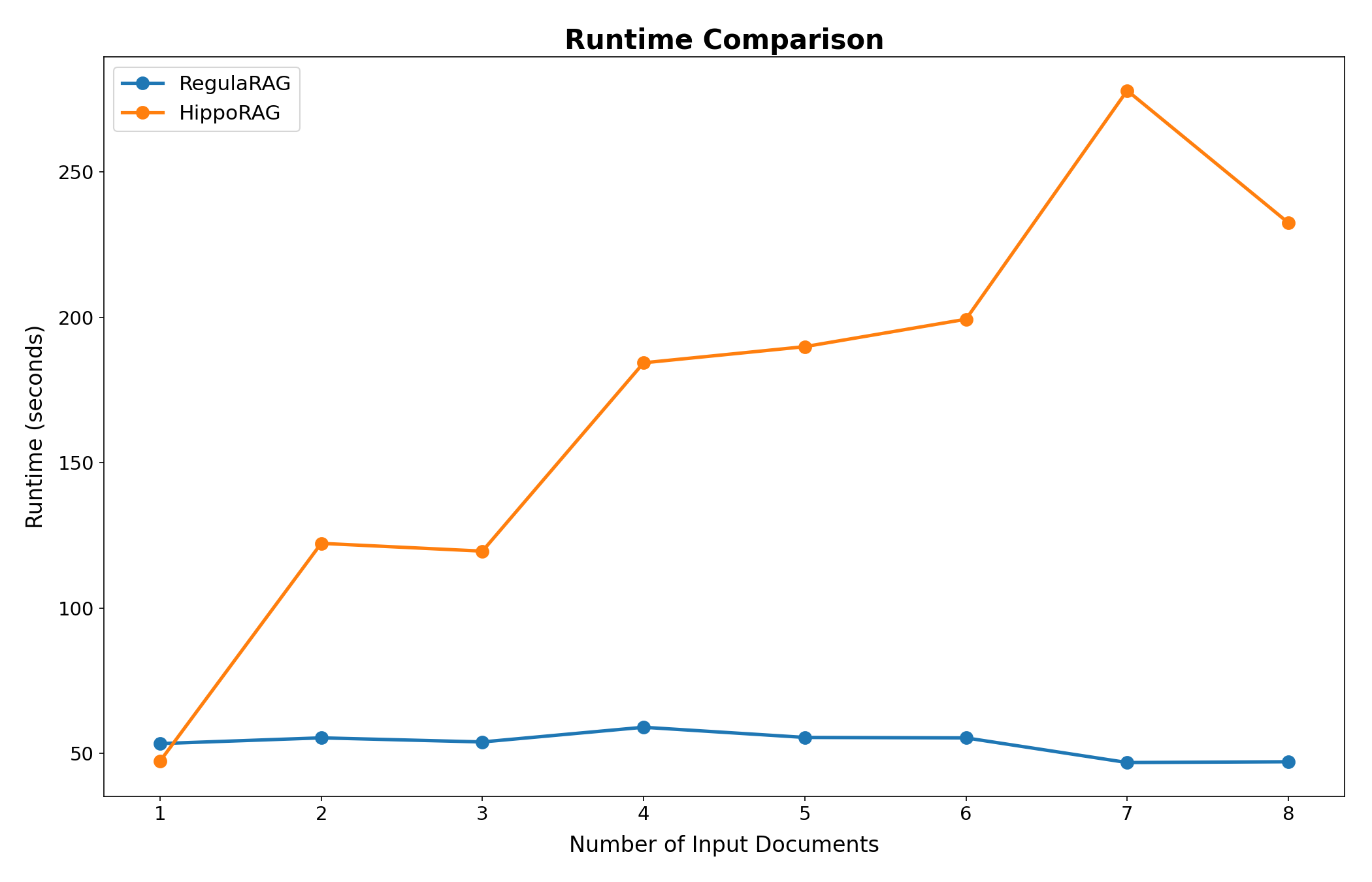}
\caption{Runtime scaling for \textsc{RegulaRAG} and \textsc{HippoRAG}.}
\label{fig:scaling_runtime}
\end{figure}

\begin{figure}[H]
\centering
\includegraphics[width=0.9\linewidth]{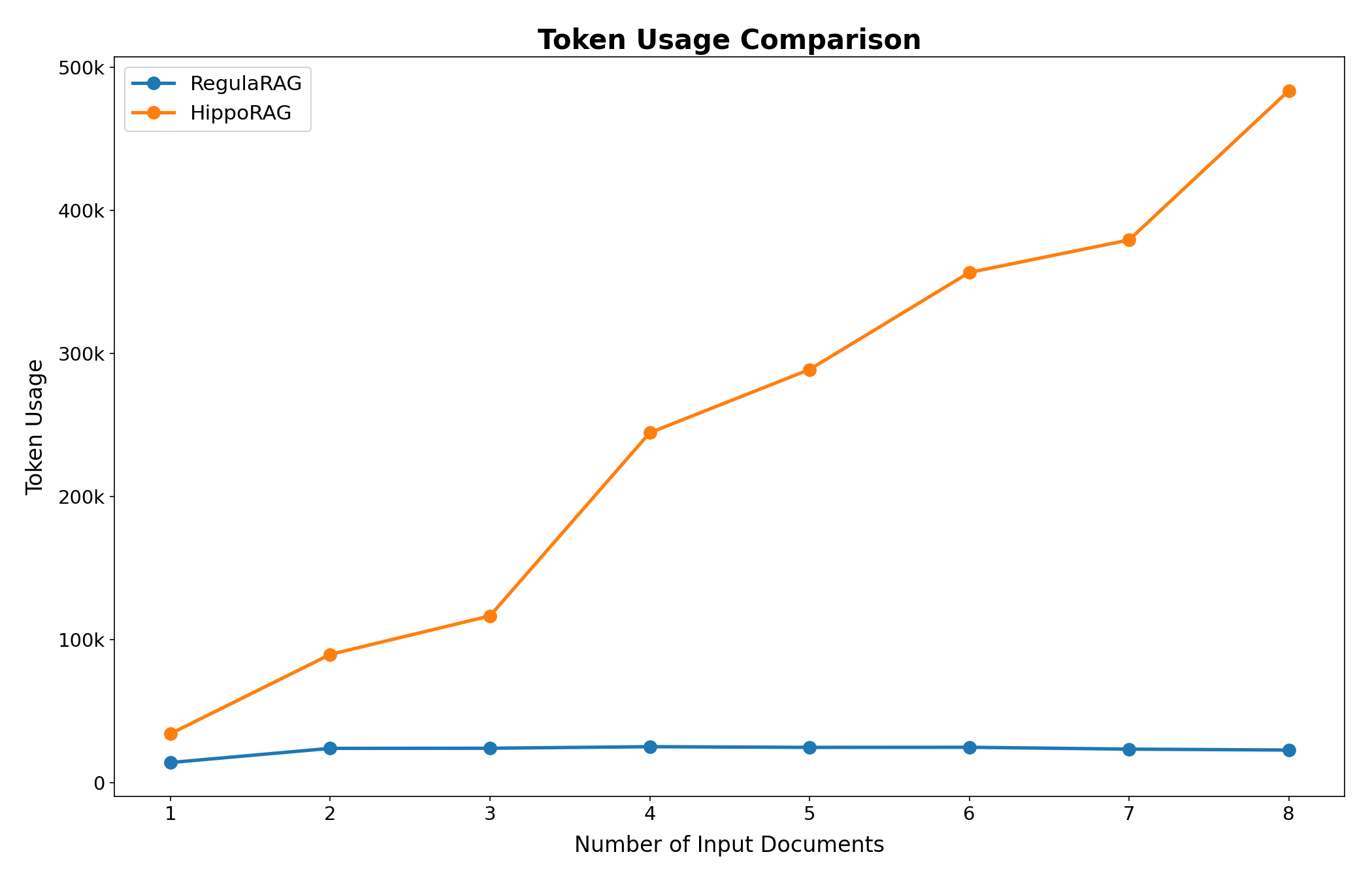}
\caption{Token usage comparison across increasing corpus sizes.}
\label{fig:scaling_tokens}
\end{figure}

Figures~\ref{fig:scaling_f1}, \ref{fig:scaling_runtime}, and \ref{fig:scaling_tokens} report F1-score, end-to-end runtime, and OpenAI token usage as the corpus grows.

\paragraph{Runtime stability.}
Across all corpus sizes, \textsc{RegulaRAG} maintains a stable runtime (47–59~s), indicating that its enrichment and reranking pipeline scales gracefully. The enrichment module has been optimized by removing deep copies, avoiding storage of enriched text, compressing metadata into integer references, and bounding BFS expansion. As shown in Figure~\ref{fig:scaling_runtime}, these optimizations enable RegulaRAG to maintain stable runtime even as the corpus grows. In contrast, \textsc{HippoRAG} runtime grows steeply, exceeding 200~s on seven or more regulations, driven by its LLM-heavy OpenIE and graph-construction passes. 
\paragraph{Token usage scaling.}
The two models consume substantially different amounts of tokens. \textsc{RegulaRAG} remains compact, with token usage fluctuating slightly around 14k–25k tokens. By contrast, \textsc{HippoRAG} grows near-exponentially, from 34k tokens on a single document to nearly 500k tokens on the full set. This growth is driven by multiple LLM-dependent stages—NER, triple extraction, graph expansion, reranking, and auxiliary calls.
\paragraph{Retrieval quality.}
\textsc{HippoRAG} delivers consistently strong accuracy (97–100\% F1), reflecting the benefits
of explicit graph-structured reasoning. \textsc{RegulaRAG}, optimized for paragraph-level
reference reconstruction, achieves high accuracy on smaller inputs
but declines as the corpus becomes increasingly heterogeneous (down to 82.35\%). This represents a
natural trade-off: RegulaRAG prioritizes computational efficiency and highly targeted retrieval,
whereas HippoRAG prioritizes robustness and semantic coverage, at the cost of substantially higher
token usage and runtime.
\subsection{Failure Mode Analysis}\label{sec:failure}

Despite RegulaRAG's strong overall performance, manual inspection of generated scenarios reveals three recurring failure patterns that affect all evaluated RAG systems to varying degrees.

\textbf{(1) Incorrect numerical value assignment.}
LLMs tend to reproduce the structural template of a scenario correctly but fail to extract and apply numerical values accurately from the regulation text.
For example, a CtoStC scenario requires the subject vehicle to travel at 20~km/h (tolerance $+0$/$-$2~km/h) before braking, derived from the regulation statement ``Tests shall be conducted with a vehicle travelling at 20, 42, and 60~km/h.''
LLMs sometimes substitute values not present in the regulation, particularly for speeds that must be read from the Maximum Relative Impact Speed table (Fig.~\ref{fig:splittable}).

\textbf{(2) Loading condition confusion.}
LLMs frequently fail to distinguish laden from unladen conditions, generating incorrect post-condition values.
The correct post-condition (e.g., the AEBS must reduce vehicle speed to 0~km/h) depends on loading state and must be read from the regulation table; confusing the two conditions yields numerically incorrect scenarios despite plausible wording.

\textbf{(3) Cosine similarity inflation.}
Because LLMs reproduce the scenario template structure faithfully, cosine similarity between generated and ground-truth scenarios is often close to 1.0 even when numerical content is wrong.
This inflates raw similarity scores and motivates the penalized metric (Algorithm~\ref{alg:penalized-scoring}), which explicitly penalises numeric and categorical discrepancies to surface these otherwise hidden errors.

RegulaRAG directly addresses failure modes (1) and (2) by ensuring the relevant table chunk is present in the retrieved context through reference-aware enrichment, thereby providing the LLM with the correct numerical grounding.
Exact numeric hallucination that occurs even when the correct table is supplied---where the LLM misreads or ignores explicit table values---is a well-known problem across regulated-domain LLM applications and remains outside the scope of this work; mitigating it through prompting strategies or constrained decoding is a planned direction for future research.

\section{Conclusion and Future Work}\label{sec:conclusion}

In this paper, we introduced RegulaRAG, a two-stage Retrieval-Augmented Generation (RAG) pipeline designed to generate regulation-compliant test scenarios from complex technical standards such as UN Regulation No.\ 152. By integrating a SmartChunking strategy with reference-aware enrichment and a semantic retrieval mechanism, RegulaRAG offers a favorable balance between accuracy, runtime, and token usage. Across multiple scenario types and language models, RegulaRAG achieves the highest average Meta-Score (82.99, 43\% above the next-best baseline NoRAG at 57.94), indicating strong and stable performance while operating at substantially lower context cost (14k–25k tokens per query) than graph-centric alternatives such as HippoRAG (up to 500k tokens on an eight-document corpus).

Future work will focus on further improving RegulaRAG, including refinements to chunk enrichment, scoring, and retrieval ranking.
Finally, we aim to generalize the pipeline to additional regulatory documents and domains, thereby broadening its applicability in compliance-critical settings.
We further envision integrating the generated scenarios directly with automotive simulation environments (e.g., CARLA, CarMaker) and real testbench infrastructure.

\subsection*{Limitations}
Several limitations of the present work merit acknowledgment.
First, RegulaRAG was developed and evaluated exclusively on UN Regulation No.~152. The core pipeline --- SmartChunking, BFS reference-closure enrichment, and semantic retrieval --- is regulation-agnostic and in principle applicable to any domain whose standards share similar structural properties: hierarchical numbered paragraphs, explicit cross-references, and tabular specifications. Domains such as pharmaceutical (e.g., ICH guidelines), aviation (e.g., DO-178C), or financial compliance standards share these structural characteristics and are plausible candidates for adaptation . Two bounded adaptation steps are required: (i)~the categorical regex family $\mathcal{R}_{\ell}$, which encodes UN-152 load-condition keywords (\textit{laden}, \textit{unladen}, \textit{MassRun}), must be replaced with domain-relevant vocabulary; the cosine-similarity backbone, numeric matching, penalty weight $\lambda$, and Meta-Score formula are domain-agnostic and transfer without modification (see Section~\ref{sec:evalc} for a full breakdown); and (ii)~the retrieval hyperparameters $(top\_k{=}30,\;chunk\_size{=}2000)$ were tuned via grid search on the CtoStC scenario family using UN-152 alone; regulations with substantially different document density, cross-reference depth, or chunk length distributions may require re-tuning of these parameters before deployment. This transfer risk may be partially bounded for $top\_k$ specifically: prior work on reader noise-sensitivity~\cite{240309040RAGGEDTowards} suggests readers with an improve-then-plateau retrieval-depth response --- a pattern GPT-4o follows alongside GPT-3.5-turbo --- degrade less when $top\_k$ is not re-tuned for a new corpus than noise-sensitive readers; this bound does not extend to $chunk\_size$, which is not examined in that work, and re-tuning both parameters remains advisable.
Second, the same subject-matter experts contributed to both dataset construction and qualitative evaluation, which introduces a potential consistency bias; role separation or an independent validation cohort would strengthen future evaluations.
Third, RegulaRAG performs reference-aware chunk enrichment and reordering but does not construct or reason over a full knowledge graph (see Section~\ref{liter} for a detailed discussion): the legal language of regulatory standards — nested clauses, conditional syntax, and domain-specific terminology — makes reliable general-purpose entity and relation extraction challenging, and KG-based pipelines incur substantially higher token and runtime costs as corpus size grows (Section~\ref{sec:scalability}). Systems that build explicit entity--relation graphs may capture deeper cross-reference semantics at the cost of this additional complexity.

Finally, owing to resource constraints, each configuration was evaluated in a single run without repeated-trial statistics. While the baseline comparison provides partial ablation evidence—R\&R+RCS isolates the contribution of BFS reference-aware enrichment and smart reranking, and NoRAG isolates the entire retrieval pipeline—a component-level ablation isolating BFS reference-closure depth, deduplication, and canonical chunk reordering individually was not conducted; these remain open empirical questions for future work.

\appendices
\section{\break User Prompts for Scenario Generation}\label{appendix:user_prompts}
In addition to the system prompt provided in Listing~\ref{lst:system_prompt}, the language 
model receives a scenario-specific user prompt. These prompts instruct the model to generate 
AEBS test scenarios in a structured format that mirrors the examples defined in 
UN Regulation No.\ 152. The three user prompts corresponding to the 
Car-to-Stationary-Car (CtoStC), Car-to-Moving-Car (CtoMoC), and Car-to-Pedestrian (CtoP) 
scenario families are listed below.

\begin{promptbox}
\textbf{User Prompt for Car-to-Stationary-Car (CtoStC)}

\textbf{CONTEXT:} \verb|{|rag\verb|_|context\verb|}|

Based on the context provided (from UN Regulation No. 152), generate all test scenarios related to Car to stationary Car scenario (CtoStC) tests for M1 vehicles. The technical service is strictly required to test any speed, therefore a complete list of all test cases (with the speed range defined in regulation tables) must be produced. The system shall be active at least within the vehicle speed range specified by Maximum Relative Impact Speed
(km/h) for M1 vehicles.

Follow this exact format for every test scenario:
1- "Write the scenario title (e.g., Test\verb|_|CtoStC\verb|_|unladen\verb|_|42)   as the first line."
2- "Immediately after the title, write the full scenario description."
3- "After each complete test case, add exactly one separator line:
---------"

Formatting rules:
- Use the provided example structure strictly.
- Do not add explanations, notes, or extra text.
- Only output the requested test scenarios and separators.
- Do not include index numbers or any other identifiers.

\textbf{Example:} \verb|{|example\verb|_|test\verb|_|case\verb|}|

[Start generation now.]
\end{promptbox}

\begin{promptbox}
\textbf{User Prompt for Car-to-Moving-Car (CtoMoC)}

\textbf{CONTEXT:}
\verb|{|rag\verb|_|context\verb|}|

Based on the context provided (from UN Regulation No. 152), generate all test scenarios
related to Car to moving Car scenario (CtoMoC) tests for M1 vehicles. The technical service
is strictly required to test any speed, therefore all speeds defined in the regulation tables
must be included. The system shall be active at least within the vehicle speed range
specified by the Maximum Relative Impact Speed (km/h) for M1 vehicles.

Follow this exact format for every test scenario:
1- "Write the scenario title (e.g., Test\verb|_|CtoMoC\verb|_|unladen\verb|_|10) as the first line."
2- "Immediately after the title, write the full scenario description."
3- "After each complete test case, add exactly one separator line:
---------"

Formatting rules:
- Use the provided example structure strictly.
- Do not add explanations, notes, or extra text.
- Only output the requested test scenarios and separators.
- Do not include index numbers or any other identifiers.

EXAMPLE:
\verb|{|example\verb|_|test\verb|_|case\verb|}|

[Start generation now.]

\end{promptbox}

\begin{promptbox}
\textbf{User Prompt for Car-to-Pedestrian (CtoP)}

\textbf{CONTEXT:} \verb|{|rag\verb|_|context\verb|}|

Based on the context provided (from UN Regulation No. 152), generate all test scenarios
related to Car to pedestrian scenario (CtoP) tests for M1 vehicles. The technical service is
strictly required to test any speed; therefore, disregard speeds mentioned in the context and
generate the complete set of speeds defined in the regulation tables. The system shall be
active at least within the vehicle speed range specified by the Maximum Impact Speed (km/h)
for M1 vehicles.

Follow this exact format for every test scenario:
1- "Write the scenario title (e.g., Test\verb|_|CtoP\verb|_|MassRun\verb|_|20) as the first line."
2- "Immediately after the title, write the full scenario description, using the exact wording
    of the provided example, adapting only the relevant parameters."
3- "After each complete test case, add exactly one separator line:
---------"

Formatting rules:
- Use the provided example structure strictly.
- Do not add explanations, notes, or extra text.
- Only output the requested test scenarios and separators.
- Do not include index numbers or any other identifiers.

\textbf{Example:} \verb|{|example\verb|_|test\verb|_|case\verb|}|

[Start generation now.]

\end{promptbox}

\section*{Acknowledgment}
This research was funded by the Federal Ministry of Research, Technology and Space (BMFTR) as part of the CeCaS project, FKZ: 16ME0800K.

\bibliographystyle{IEEEtran}
\bibliography{My_Library}

\begin{IEEEbiography}[{\includegraphics[width=1in,height=1.25in,clip,keepaspectratio]{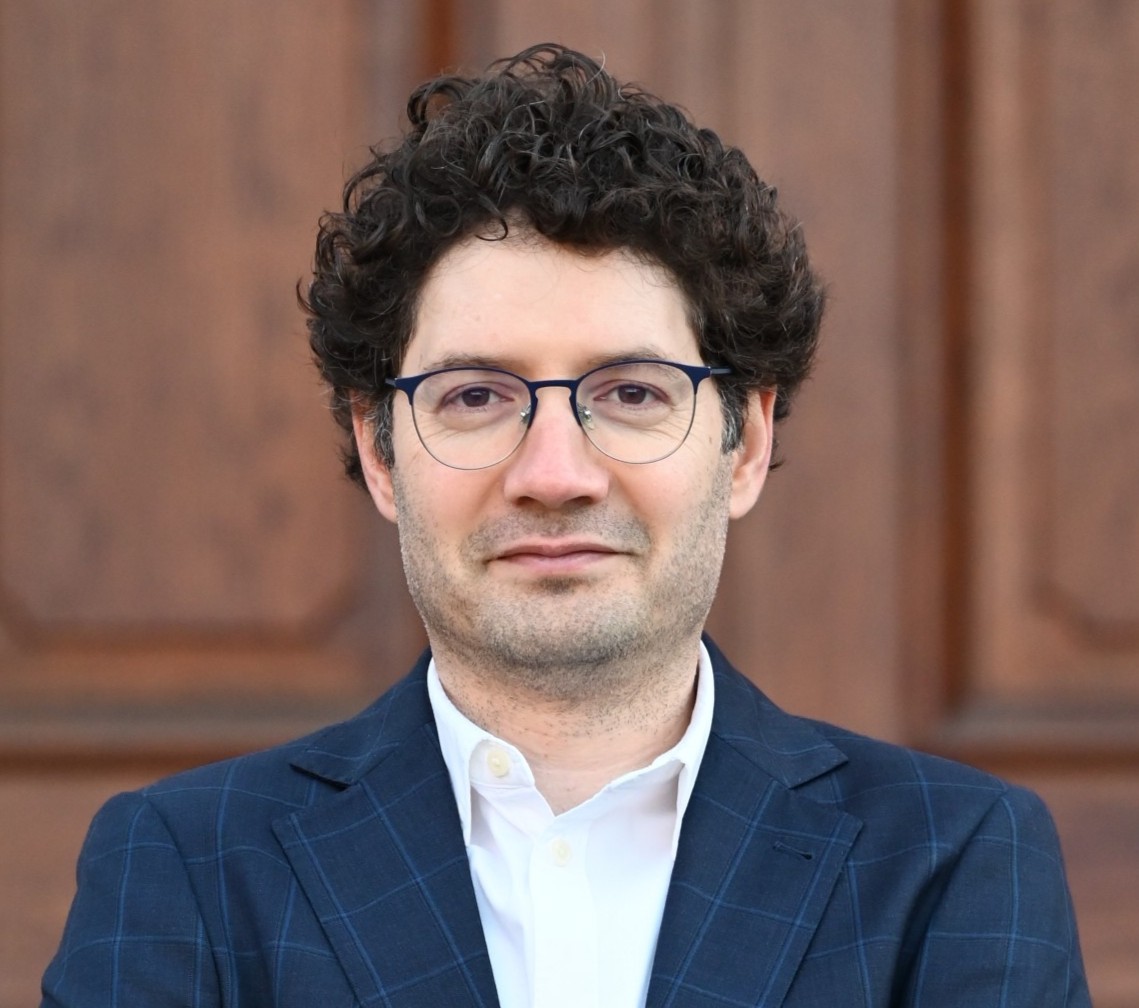}}]{VAHID ZOLFAGHARI} received the M.Sc. degree in computer science from the Amirkabir University of Technology. He is currently pursuing the Ph.D. degree with the Technical University of Munich. He is part of the Central Car Server (CeCaS) Project. Previously, he was a Senior Test Engineer in network security and later a Research Assistant in Italy, contributing to projects in IoT security, robotic systems, and neurorobotics. He has co-authored several publications on generative AI and automotive system design. His research focuses on the application of large language models and retrieval-augmented generation systems for automotive software engineering.
\end{IEEEbiography}
\begin{IEEEbiography}[{\includegraphics[width=1in,height=1.25in,clip,keepaspectratio]{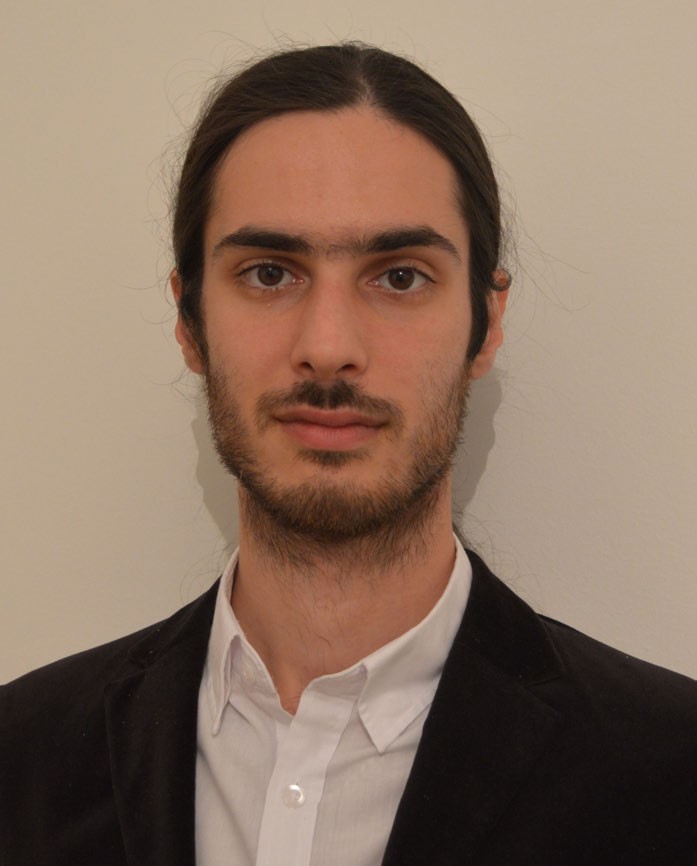}}]{Nenad Petrovic} was born in Pirot, Serbia,
in 1992. He received the bachelor’ degree in
computer science and informatics from the Facultyof Electronic Engineering, University of Nis, Nis, Serbia, in 2015, the Laurea Magistrale degree in computer engineering from the Politecnico di Milano, Milan, Italy, and the Ph.D. degree in computing and electrical engineering from the Faculty of Electronic Engineering. He was a Teaching Assistant. He is currently a Postdoctoral Researcher and a Scientific Coordinator/Supervisor with the Chair of
Robotics, Artificial Intelligence and Real-Time Systems, Technical University of Munich (TUM), focused on automotive-related projects in area of GenAI-driven software development. He is the author or co-author of more than 200 scientific publications. His main areas of interests include
model-driven software engineering, semantic technology, the Internet of Things (IoT), and generative artificial intelligence (GenAI).
\end{IEEEbiography}
\begin{IEEEbiography}[{\includegraphics[width=1in,height=1.25in,clip,keepaspectratio]{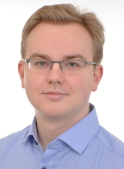}}]{ANDRÉ SCHAMSCHURKO} received the Dipl.Inf. degree in computer science from TU Dresden,
in 2023. He is currently pursuing the Ph.D. degree with the Chair of Robotics, Artificial Intelligence and Real-Time Systems, Technical University of Munich. The primary focus of his research is
on natural language processing and generative AI models, with a particular interests in reducing their hallucinations.
\end{IEEEbiography}

\begin{IEEEbiography}[{\includegraphics[width=1in,height=1.25in,clip,keepaspectratio]{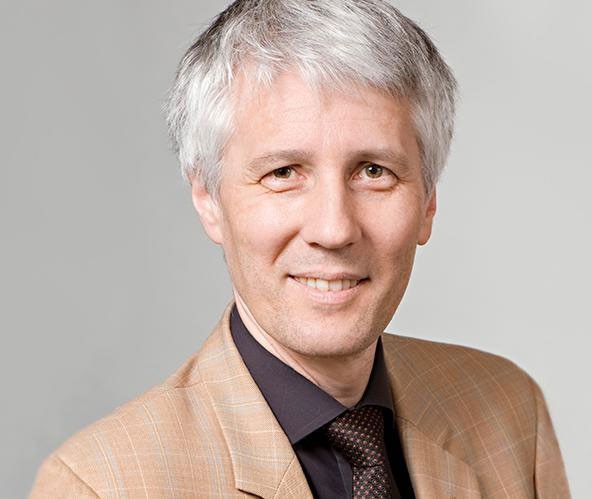}}]{ALOIS KNOLL} (Fellow, IEEE) received the M.Sc.
degree in electrical/communications engineering
from the University of Stuttgart, Stuttgart, Germany, in 1985, and the Ph.D. degree (summa cum
laude) in computer science from the Technical
University of Berlin (TU Berlin), Berlin, Germany,
in 1988. He was on the Faculty of the Computer Science Department, TU Berlin, until 1993.
He joined the University of Bielefeld, Bielefeld, Germany, as a Full Professor, where he was the Director of the Technical Informatics Research Group, until 2001. Since 2001, he has been a Professor with the Department of Informatics, Technical
University of Munich (TUM), Munich, Germany. He was on the board of
directors of the Central Institute of Medical Technology at TUM (IMETUM).
From 2004 to 2006, he was the Executive Director of the Institute of
Computer Science, TUM. From 2007 to 2009, he was a member of the EU’s
highest advisory board on information technology, ISTAG, the Information
Society Technology Advisory Group, and its subgroup on future and
emerging technologies (FET). In this capacity, he was actively involved in
developing the concept of EU’s FET flagship projects. His research interests
include cognitive, medical, and sensor-based robotics; multi-agent systems;
data fusion; adaptive systems; multimedia information retrieval; modeldriven development of embedded systems, with applications to automotive
software and electric transportation; and simulation systems for robotics and
traffic.
\end{IEEEbiography}

\end{document}